\documentclass[letterpaper, 10 pt, conference]{ieeeconf}

\IEEEoverridecommandlockouts

\usepackage{graphicx} 
\usepackage{epstopdf} 
\usepackage{amsmath} 
\usepackage{amssymb}  
\usepackage{algorithm}
\usepackage{algpseudocode}
\usepackage{bm}

\newcommand{\vc}[1]{{\bm #1}}

\IEEEoverridecommandlockouts
\newcommand{\headertext}{USC TECH REPORT - SPRING 2018}
\usepackage{fancyhdr}
\fancypagestyle{pageStyleOne}{%
    \fancyhf{}
    \fancyhead[C]{\headertext}
}

\usepackage{cite}

\title{\LARGE \bf
Planning Safe Paths through Hazardous Environments*
}

\author{Chris Denniston$^{1 \dagger}$ and Thomas R. Krogstad$^{2 \dagger}$ and Stephanie Kemna$^{1}$ and Gaurav S. Sukhatme$^{1}$
\thanks{*This work was supported by ONR and the Norwegian Defence Research Establishment.}
\thanks{$^{\dagger}$ equal contributions}
\thanks{$^{1}$Chris Denniston, Stephanie Kemna, and Gaurav S. Sukhatme are with Department of Computer Science,
        University of Southern California, Los Angeles, USA
        {\tt\small \{cdennist,kemna,gaurav\}@usc.edu}}%
\thanks{$^{2}$Thomas R. Krogstad is with Norwegian Defence Research Establishment (FFI), P.O. Box 25, N-2027 Kjeller, Norway.
        Norway
        {\tt\small Thomas.Krogstad@ffi.no}}%
}

\begin{document}

\maketitle

\thispagestyle{pageStyleOne}
\pagestyle{fancy}

\begin{abstract}
Autonomous underwater vehicles (AUVs) are robotic platforms that are commonly used to map the sea floor, for example for benthic surveys or for naval mine countermeasures (MCM) operations.
AUVs create an acoustic image of the survey area, such that objects on the seabed can be identified and, in the case of MCM, mines can be found and disposed of.
The common method for creating such seabed maps is to run a lawnmower survey, which is a standard method in coverage path planning. 
We are interested in exploring alternate techniques for surveying areas of interest, in order to reduce mission time or assess feasible actions, such as finding a safe path through a hazardous region.
In this paper, we use Gaussian Process regression to build models of seabed complexity data, obtained through lawnmower surveys.
We evaluate several commonly used kernels to assess their modeling performance, which includes modeling discontinuities in the data. Our results show that an additive Mat\'ern kernel is most suitable for modeling seabed complexity data.
On top of the GP model, we use adaptations of two standard path planning methods, A* and RRT*, to find safe paths for marine vessels through the modeled areas.
We evaluate the planned paths and also run a vehicle dynamics simulator to assess potential performance by a marine vessel.
\end{abstract}

\section{Introduction}

Autonomous underwater vehicles (AUVs) equipped with side-looking sonars have become an increasingly important asset in naval mine countermeasures (MCM) operations, by lowering the cost of operations and reducing the risk to the personnel on the vessels. 
AUVs are typically deployed for the mine detection and classification phase of the MCM operations. 
For this phase, the vehicles are programmed to run pre-planned lawnmower surveys, i.e. Boustrophedon motions, to create a full coverage high-resolution acoustic image of the survey area.

\begin{figure}[tbh]
  \centering
  \includegraphics[width=\columnwidth]{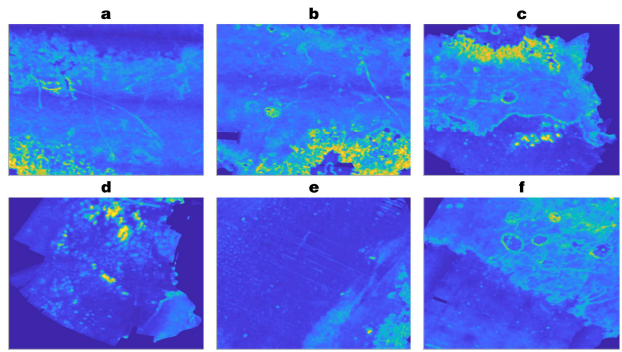}
  \caption{Seabed complexity data gathered in the field by the Hugin AUV, showing six examples or scenarios. Color ranges from low complexity (deep blue) to high complexity (yellow).}
	\label{fig:scenarios}
\end{figure}

The use of high-resolution side-scan sonar (SSS) and synthetic aperture sonar (SAS) together with new methods to analyze the data have given the AUVs a new ability to evaluate the complexity of the seabed and to predict the MCM performance \cite{Geilhufe2016}.
For example, AUVs can estimate the probability of detecting a mine-like object in a given area by considering seabed complexity.
Figure~\ref{fig:scenarios} shows six examples of seabed complexity data collected during prior lawnmower surveys with a Kongsberg HUGIN 1000 class AUV \cite{Hagen2003}.

We are interested in exploring in how far we can develop alternative methods of surveying areas of interest, for example to decrease overall mission time, or to more quickly assess feasible operations such as traversal through the area.
In order to do so, we want the AUV to create on-board models of the area of interest based on sensor data.
Gaussian Process (GP) regression is one technique that can be used to create a model that interpolates between sampled points. It also estimates uncertainty in the model, which is useful for deciding where to focus sampling efforts. 
In this paper we evaluate the GP regression performance for MCM scenarios by testing different GP kernels, i.e. its covariance functions.
Furthermore, we investigate the potential of existing path planning methods that can run on these models to find paths for safe traversal by marine vessels.

\subsection{Related Work}

GP regression is also known as Kriging in the field of spatial statistics, where it is used primarily as an interpolation technique.
Geospatial Kriging assumes there is some spatially related component that represents a trend or a random error term~\cite{Chang2006}, based on what is informally called Tobler's first law of geography: ``Everything is related to everything else, but near things are more related than distant things.''~\cite{Tobler1970}.
Section~\ref{sec:gp} will explain the basics of GP regression.

A Gaussian process has a kernel that specifies the extent to which nearby points are inferred based on sampled points.
Most related works in environmental modeling and adaptive sampling use the squared exponential (SE) kernel~\cite{Krause2008, Low2008, Yang2013, Hollinger2013}. This is a smooth kernel, as further defined in section~\ref{sub:choose_kernel}.
The smoothness assumptions are valid for the modeling of environmental characteristics, such as water temperature or algae blooms, but may not hold for seabed complexity, where discrete boundaries may be present.
Vasudevan et al.~\cite{Vasudevan2009} evaluated the squared exponential and neural network (NN) kernels for modeling terrain data.
Their approach uses a localized neural network kernel because of the large number of discontinuities in terrain data.
The authors found that the neural network kernel outperforms the squared exponential kernel for that task.
Therefore, we use the NN kernel as one of the kernels in our evaluation.

\subsubsection*{Path Planning on GP Models and for MCM Operations}

Yang et al.~\cite{Yang2013} used a GP to create an occupancy map for unmanned aerial vehicles, and apply an RRT-based planner for finding safe paths through an obstacle-rich environment in 3D.
In this paper, we extend their result by considering the environment's complexity as observed by the sensor, by comparing to A* path planning, and by comparing the effect of using different covariance functions in the GP regression model.

Hollinger \& Sukhatme~\cite{Hollinger2014} presented an RRT-based path planning approach in an exploration of different sampling-based methods for off-line path planning on top of GP models.
In this paper, we similarly consider a sampling-based planner, RRT*, but also incorporate kinematic constraints into the path planning.

The current use of AUVs in typical MCM operations is to provide a detailed high-resolution sonar map of the near-shore seabed. 
Most path planning algorithms for MCM operations focus on planning for full coverage~\cite{Galceran2013,Paull2014,Krogstad2014, Williams2010}.
We are interested in the problem of planning a safe path through an area that may contain mines. We assume that a coverage method has been used for creating a map of the area.

One approach that investigated path planning for a safe path through a mined area was presented by Bekker and Schmid~\cite{Bekker2006}.
The authors use Dijkstra's algorithm and genetic algorithms to find a path of minimum risk, or a path of minimum length with acceptable risk, through a mined area. 
If no safe path is found, a path which requires a minimum number of mines to be removed is given.
The area of operation is assumed to be perfectly known and all mines are assumed detected. The authors find that these techniques provide a basis for finding safe paths through risky areas.
Our approach extends theirs by considering the through-the-sensor estimated MCM performance and considering paths that respect the kinematic constraints of the vessel in question.

\subsection{Contribution and Overview}
The contributions of this paper are three-fold:
\begin{enumerate}
\item We evaluate different kernels for Gaussian Process regression to find the one that best models seabed complexity data, which may contain discontinuities,
\item We evaluate the performance of two path planning methods, RRT* and A*, for finding a safe path through an area, assuming a GP model has been created.
\item We evaluate the feasibility of planned paths by running a vehicle dynamics simulator, and assess the performance difference.
\end{enumerate}

We find that a kernel using a linear mixture of two Mat\'ern class kernels outperforms all standard kernels, and that Mat\'ern class kernels outperform the commonly used Squared Exponential kernel.
Using a GP model with the additive Mat\'ern kernel, we evaluate the performance of an A* path planner, and RRT* with Dubins constraints.
Our results show that the performance of the path planning algorithms is scenario dependent.
We consider both path length and path complexity in our path cost, and the choice of the weighting affects the quality of the planned paths. 
When we run the vehicle dynamics simulator, we see that the executed paths tend to have a greater length and increase path cost.
Because the RRT* algorithm better incorporates vehicle dynamics, the vehicle dynamics simulator is better able to follow the RRT* planned paths than the A* planned paths.

\section{Gaussian Process Regression for Field Modeling}\label{sec:gp}

Gaussian process regression is used to approximate some unknown function $f$ from its known outputs. 
It does this by approximating any points $\mathbf{x}$ in the input space by Gaussians, based on measurements $y$ taken at that input location and nearby measurements.
In how far other measurements affect the estimate at a certain input location is determined by the covariance function, or kernel $k(\cdot, \cdot)$, that is used for the regression.

Function values $y_*$ at any input location $\mathbf{x}_*$ are approximated by a Gaussian distribution:
\begin{equation}\label{eq:gp_inference}
 y_* | y \sim \mathcal{N} \left(K_* K^{-1} y,~ K_{**} - K_* K^{-1} K^T_*A \right) 
\end{equation}
where
\begin{quote}
  $y$ is training output,\\  
  $y_*$ is test output,\\
  $K$ is $k(\mathbf{x},\mathbf{x}),$\\
  $K_*$ is $k(\mathbf{x},\mathbf{x_*}),$\\
  $K_{**}$ is $k(\mathbf{x_*},\mathbf{x_*}),$\\
  $\mathbf{x}$ is training input, and\\
  $\mathbf{x_*}$ is test input.\\
\end{quote}
The choice of kernel $k(\cdot, \cdot)$ is a principal architectural choice for GP regression, which represents underlying assumptions about the data. 
Most kernels have some parameters that need to be set, though these typically can be estimated from training data using maximum likelihood estimation. 
These kernel parameters are known as the hyperparameters of the GP. Typical hyperparameters include the kernel's length scale, i.e. the extent to which neighboring points are updated based on a measurement, and the kernel's signal variance, also known as the amplitude~\cite{Rasmussen2006}.

\subsection{GP Kernel Choice}\label{sub:choose_kernel}

We evaluate five different kernels: 
the Squared Exponential (SE) kernel, 
three versions of the Mat{\'e}rn kernel, 
and the Neural Network kernel.
We briefly go over the equations and parameters per type of kernel. 
Then we explain how we tested the kernels on real data, and provide the results and our choice of kernel.

For equations \ref{eq:se_kernel} through \ref{eq:matern_mixture}:
\begin{quote}
$\mathbf{x}$, $\mathbf{x}'$ are training inputs,\\
$\sigma^2$ represents the signal variance (amplitude),\\
$\ell$ represents the length scale.
\end{quote}
where $\sigma^2$ and $\ell$ are both hyperparameters for the kernels.

\subsubsection{Squared Exponential Kernel}
The Squared Exponential kernel is defined as \cite{Rasmussen2006}:
\begin{equation}\label{eq:se_kernel}
	k(\mathbf{x}, \mathbf{x}') =  \sigma ^2 \exp \left[ -\frac{1}{2} \left( \frac{\mathbf{x} - \mathbf{x}'}{\ell} \right) ^2 \right]
\end{equation}
The squared exponential kernel is a widely used kernel in GP regression, providing a baseline to compare other kernels to.
The kernel is stationary, isotropic, infinitely differentiable and very smooth~\cite{Rasmussen2006,Vasudevan2009}.

\subsubsection{Mat\'ern Kernel}
The Mat\'ern class is a class of kernels defined as \cite{Rasmussen2006}:
\begin{equation}\label{eq:mat_kernel}
	k(\mathbf{x}, \mathbf{x}') = \sigma ^2  \frac{2^{1-v}}{\Gamma(v)} \left( \frac{\sqrt[]{2v} |\mathbf{x} - \mathbf{x}'|}{\ell} \right)^v K_v\left( \frac{\sqrt[]{2v} |\mathbf{x}-\mathbf{x}'|}{\ell} \right)
\end{equation} 
 where $v$ is a chosen parameter; $K_v$ is a modified Bessel function, $\Gamma$ is the Gamma distribution \cite{Rasmussen2006}.
The Mat\'ern class is characterized by the parameter $v$, which defines that the function is $k$ times differentiable for $v > k$, as opposed to infinitely differentiable for the squared exponential kernel.
The Mat\'ern kernel is the same as the SE kernel when $v \rightarrow \inf$~\cite{Rasmussen2006}.
This difference is key in explaining the difference in performance between the two kernel classes.
Stein has suggested that ``[Infinite differentiability] would normally be considered unrealistic for a physical process'' and offers the Mat\'ern class as an alternative~\cite{Stein1999}.
Common choices for the parameter $v$ are $v=5/2$ and $v=3/2$~\cite{Rasmussen2006}.
These two parameters have been used in this paper, where we replace the parameter $v$ with $p$ such that $v=p/2$, and $p = \{3, 5\}$.

\subsubsection{Neural Network Kernel}
The neural network kernel is defined as \cite{Rasmussen2006}:
\begin{equation}\label{eq:nn_kernel}
	k(\mathbf{x},\mathbf{x}') = \sigma ^ 2 \arcsin \left( \frac{\mathbf{x}^T \ell \mathbf{x}'}{\sqrt[]{(1+\mathbf{x}^T \ell \mathbf{x})(1+\mathbf{x}'^T \ell \mathbf{x}')}} \right)
\end{equation}
The neural network kernel is similar to a single hidden layer neural network with infinitely many hidden nodes and a sigmoid transfer function~\cite{Vasudevan2009}.

\subsubsection{Mat\'ern Additive Model}
We also include a model that is a linear combination of the two common parameter choices for the Mat\'ern kernel:
\begin{equation}\label{eq:matern_mixture}
	k(\mathbf{x},\mathbf{x}') = \alpha k_3(\mathbf{x},\mathbf{x}') + \beta k_5(\mathbf{x},\mathbf{x}')
\end{equation}
where $\alpha$ and $\beta$ are scale hyperparameters, $k_3$ is the Mat\'ern kernel with $p=3$ and $k_5$ is the Mat\'ern kernel with $p=5$. $\alpha$ and $\beta$ are initialized to $1$, and are automatically updated during hyperparameter optimization.

\subsection{Kernel Testing Set-up}

To choose the kernel best suited to our application, we evaluate kernel performance over prior collected field data. 
We have six scenarios from real data, shown in Figure~\ref{fig:scenarios}.
The data consists of sea bed complexity estimates generated from high-resolution SAS images collected by a HUGIN 1000 class AUV. 
To test modeling performance per kernel, we run lawnmower surveys.
Lawnmower surveys, also called Boustrophedon motions, are a typical method for coverage planning~\cite{Choset2000}.
This is also the standard method used by AUVs in MCM operations for mapping an area of the seabed.

The GP's kernels are compared in terms of the root mean squared error (RMSE) between the GP model and the ground truth data.
For each kernel, the parameters are initialized to a length scale of $10\,m$ and a signal noise of $0.1$. These are chosen based on expert knowledge, and are re-estimated from the sampled data after the first lawnmower turn.
The hyperparameters are re-estimated on a randomly selected half of the collected data after the first, and after every fourth turn, to consider newly gathered data. 
We limit the hyperparameters from changing more than $200\%$ in one update, to avoid erroneous updates resulting from local optima in the optimization routine. 

We run all simulations in Matlab, using the GPML libraries~\cite{gpml}, and simulate an AUV with a maximum sonar range of $20\,m$.
Sonar occlusion caused by the AUV is simulated such that the minimum sensor range is $10\,m$, i.e. there is $10\,m$ occlusion on each side of the vehicle.
Sensor noise is added with a uniform distribution $\pm 0.01\%$ of the maximum value in a field. 

For the neural network kernel, as per equation~\ref{eq:nn_kernel}, we have adopted a technique from Vasudevan et al.~\cite{Vasudevan2009}. 
They use a local approximation method akin to only using the $k$ nearest neighbors to the location to evaluate the model.
In our experiments $k=10$ is used, which is of a similar scale to $k=100$ as used by Vasudevan et al.~\cite{Vasudevan2009}, given the smaller scale of our scenarios.

\subsection{Kernel Testing Results}

\begin{figure}[!t]
  \centering
  \includegraphics[width=\columnwidth]{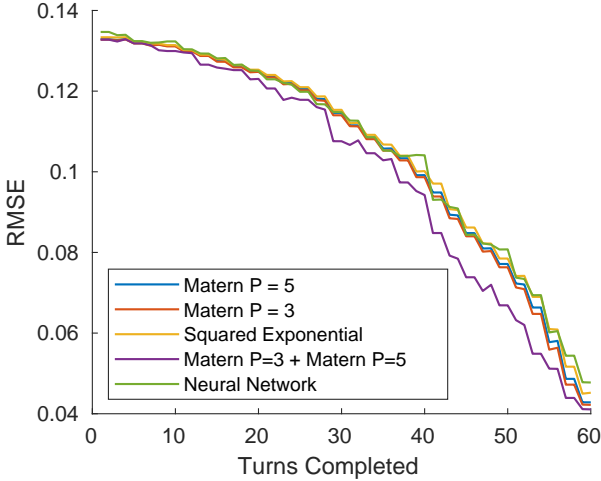} 
  \caption{RMSE curve displaying GP modeling performance given the listed kernels, tested on scenario b)  (Figure~\ref{fig:scenarios}).} 
  \label{fig:gp_rmse}
\end{figure}

Figure \ref{fig:gp_rmse} shows the RMSE versus the number of lawnmower turns completed for scenario b, Fig.~\ref{fig:scenarios}.
The RMSE curve represents the predictive power of the GP as the survey progresses.
As can be seen in Figure~\ref{fig:gp_rmse}, the additive Mat\'ern kernel performs best.
We show the results for only one scenario here. 
These results are representative for all scenarios, and the full set of results are included in appendix section~\ref{app:kernel}, page~\pageref{app:kernel}.
In all other scenarios, besides d, the Mat\'ern Additive kernel also outperforms all the other kernels. 
Furthermore, in 5 out of 6 scenarios (a, b, c, e, f) the Mat\'ern class of kernels outperforms the Squared Exponential kernel. 
The Neural Network kernel performs the most poorly in all scenarios besides b and d.

In general, each scenario will have a different overall curve due to the different structure and order of features in the scenario. 
Scenarios with more features have a higher starting RMSE with a more drastic curve downwards as the model receives more examples. 
In these scenarios the Mat\'ern Additive Model improves its model significantly faster than the other kernels, and is able to extrapolate more on partially estimated fields.

\begin{figure}[!b]
  \centering
  \includegraphics[width=\columnwidth]{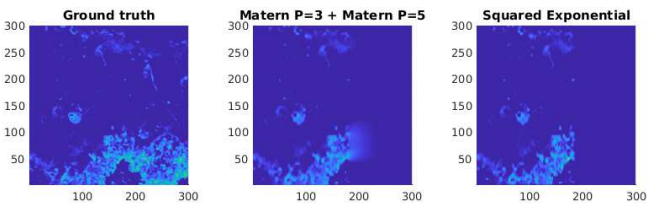}
  \caption{Example inference from GP on partially completed survey}
  \label{fig:gp_inf}
\end{figure}

Figure \ref{fig:gp_inf} indicates the GP's extrapolation ability for the additive Mat\'ern and SE kernels, which shows the predicted field after 36 lawnmower turns.
The graph demonstrates that the Mat\'ern kernel has better extrapolation power than the SE kernel.
Based on both the RMSE performance and the ability to extrapolate, we choose to use the additive Mat\'ern kernel for GP regression to model seabed complexity data.

\section{Planning Safe Paths for Marine Vessels}

In this section we present our path planning approach.
We assume that an initial survey of the area has been performed, using either a pre-planned or adaptively executed lawnmower pattern.
This survey is used to create a GP model of the seabed complexity, to assess the potential MCM performance in the area of interest.
Our path planning objective is to calculate a safe path through the surveyed area. 
A safe path in this context is a path of minimum length, which avoids collisions and very shallow waters, and which has acceptably low risk. 
Low risk is assessed by proxy as the seabed complexity; a lower seabed complexity indicates a higher probability of detecting objects and a lower risk.
We compare two standard methods for path planning; RRT* with kinematic constraints and A*.


\subsection{A* Path Planner}\label{sub:astar}
The A*-algorithm is a search algorithm that was introduced as an extension of Dijkstra's algorithm for graph search~\cite{Hart1968}. 
The algorithm uses a heuristic $h(n)$, which estimates the cost from a node to the goal, and the exact cost $g(n)$ from the starting location to any node, to prioritize which nodes to expand first. This is done by selecting the node with minimum value
\begin{equation}
f(n) = h(n) + g(n)
\end{equation}
It will thus avoid expanding paths that are already too expensive, and will more quickly reach a feasible path than Dijkstra's algorithm.
The heuristic function needs to be admissible, i.e. $h(n) \leq h^*(n)$, where $h^*(n)$ is the true cost to reach the goal state from the current state $n$.
If so, A* will reach an optimal solution. In this paper the Euclidean distance is used as the heuristic, i.e. $h(n_i) = || p_g - p_i ||_2$ where $p_g$ is the position of the goal and $p_i$ is the position of node $i$.
To avoid that A* chooses through very narrow passages, that may be infeasible for an aquatic robot, we 
consider the cost of points $5\,m$ around the point for consideration.

We define our cost function as
\begin{equation}
g(n) = l(n) + \alpha \chi(n)
\end{equation}
where $l(n)$ 
is the path length from the start node to the current node $n$, $\chi(n)$ is the accumulated complexity along the path,
and $\alpha$ is a weight used to scale complexity cost against the length cost. 
We explore the choice of $\alpha$ in section~\ref{sub:pp_results}.

\subsection{RRT* with Kinematic Constraints}\label{sub:rrt}

Rapidly-exploring random trees (RRT) were first introduced by LaValle~\cite{Lavalle1998}. 
RRT was designed as a randomized data structure and algorithm for efficiently planning in high-dimensional spaces. 
The main advantage is in its ability to quickly explore a given configuration space or state space at low computational cost.
The algorithm has been shown to be probabilistically complete, i.e. given that a solution exists, the algorithm will find it given infinite time.
The RRT algorithm works by building a path towards the goal state by randomly selecting a configuration or new state. The nearest node in the tree is extended towards the new configuration and, given that a feasible collision free path exists, the new configuration is added to the tree.

An optimal version of the rapidly-exploring random tree, RRT*, was introduced by Karaman \& Frazzoli~\cite{Karaman2011}.
RRT* mainly differs from RRT in the way the tree is extended towards the random sample configuration, by considering not only the nearest node, but a set of nearest nodes within a ball in the configuration space. 
In addition, once the new configuration has been added to the tree, the nodes in the set of nearest nodes are rewired to this new node if it results in a lower path cost.

In our approach, we plan for a vehicle with non-holonomic constraints, moving at constant or slowly varying positive forward speed. 
To this end, we follow the approach in Karaman et al.~\cite{Karaman2011a}, which uses a \emph{Steering} function that assumes a Dubins vehicle model~\cite{Dubins1957} to generate feasible trajectories between nodes. 
The non-holonomic constraint additionally affects the choice of cost function, because we now use the Dubins path length when calculating the cost to move between nodes. 
The Dubins paths  are the set of admissible paths generated assuming a dynamic model of a unicycle vehicle moving at a constant forward speed: 
\begin{subequations}
\begin{align}
\dot{x} &= u \cos(\psi)\\
\dot{y} &= u \sin(\psi)\\
\dot{\psi} &= r,
\end{align}
\end{subequations}
where $u$ is the forward speed, $\psi$ is the heading, $(x,y)$ is the position of the vehicle in $\mathbb{R}^2$ and $r$ is the angular velocity. 
$r \in \left[ -r_{\text{max}}, r_{\text{max}} \right]$ and $r_{\text{max}} \triangleq u/\rho_{\text{min}}$, where $\rho_{\text{min}}$ is the minimum turning radius.

Karaman \& Frazzoli~\cite{Karaman2011} define path cost $c(n_i, n_j)$ as the Dubins path length between the two nodes. 
As in the case of the A* algorithm in \ref{sub:astar}, we define a cost function which penalizes both the path length as well as the accumulated path complexity along the path:
\begin{equation}
c(n_i, n_j) = l_d(n_i,n_j) + \alpha \chi(n_i, n_j),
\end{equation}
where $l_d(n_i,n_j)$ is the length of the Dubins curve connecting $n_i$ and $n_j$, $\chi(n_i, n_j)$ is the accumulated complexity along the path connecting the nodes, and $\alpha$ is a weight term. 
By increasing $\alpha$, the path planning algorithm may find paths further from the minimal length path that are considered safer.

Both of the path planners were implemented in Matlab.
To evaluate path planner performance, each planner was run on the same scenarios used to select the GP kernel, shown in Figure~\ref{fig:scenarios}. 
Each scenario consists of seabed complexity data, which is generated from high-resolution sonar data collected by a Kongsberg HUGIN 1000 class vehicle. Geilhufe~\cite{Geilhufe2016} describe a method for the calculation of seabed complexity data from SAS data.
Using the procedure described in \ref{sec:gp}, we sample the complexity data as measurements along a lawnmower pattern to generate a GP regression model.
The GP model is used to predict a complexity field, which is used for the path planners' cost function $\chi(n)$.
In addition, we create an obstacle map by thresholding on the predicted complexity field for a maximum allowed value.
For our seabed complexity, where values are normalized from $0$ to $1$, we use a maximum allowed value threshold of $0.9$.
For each scenario we run the planners with five different weights $\alpha = \{0,0.25,0.75,2,1000\}$, which weight the importance of the complexity cost functions.

\subsubsection*{Vehicle Dynamics Simulation}
To better evaluate the performance of the A* and RRT* algorithms, the planned path was also used as a reference trajectory in a dynamic model of an under-actuated marine surface vessel. 
The surface vessel is modeled as~\cite{Fossen2011}: 
\begin{equation}
 \dot{\vc{\eta}} = \begin{bmatrix}
 \vc{R}(\psi) & 0 \\  0 & 1
 \end{bmatrix} \vc{\nu}, ~
 \vc{M}\dot{\vc{\nu}} + \vc{C}(\vc{\nu})\vc{\nu} + \vc{D}\vc{\nu} = \vc{B}\vc{\tau}
\end{equation}
where $\vc{\eta} = [x,y,\psi]^T$, $\vc{\nu}=[u,v,r]$, $(x,y)$ is the position, $\psi$ heading, $\vc{R}(\psi) \in SO(2)$ is a rotation matrix about the z-axis,  $u$ and $v$ are surge and sway velocity respectively, and $r$ the angular velocity.
$\vc{M} = \vc{M}^T > 0$ is a symmetric positive definite inertia matrix including added mass, $\vc{D} > 0$ is the hydrodynamic damping matrix, and $\vc{B}$ is the actuator configuration matrix. 
$\vc{\tau}$ is the actuator input, with components surge thrust $T_u$ and rudder angle $\delta$. Finally, $C(\nu)$ is the skew-symmetric coriolis matrix. 
The vessel is controlled by a line-of-sight guidance controller which calculates the desired heading angles $\psi_d$ as
\begin{equation}
 \psi_d \triangleq \arctan(-e, \Delta)
\end{equation}
where $e$ is the cross-track error and $\Delta$ is the look-ahead distance.
We use a look-ahead distance of $20\,m$ for our simulations.
The feedback linearizing controller presented in Moe et. al~\cite{Moe2014} is used to track the desired heading $\psi_d$.

From simulating the vehicle movements using this dynamic model, we obtained a path as it would be executed by a marine surface vessel. 
In the analysis of the path planning algorithms, we compare the cost and length for each path for both the planned path and the path as executed by the simulated vessel.

\subsection{Path Planning Results}

\begin{figure}[!b]
	\centering
	\includegraphics[width=.95\columnwidth]{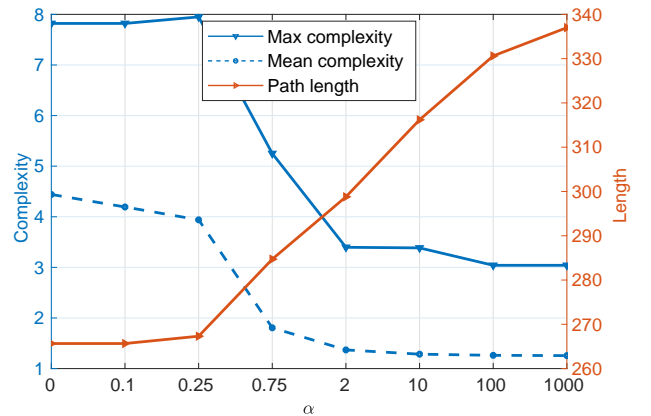} 
	\caption{Path length and mean and max complexity for different values of $\alpha$.}
	\label{fig:choice_alpha}
\end{figure}

We evaluate the different path planners in terms of the path cost, which includes path complexity and path length. 
Figure~\ref{fig:choice_alpha} shows the mean and max path complexity (left axis) for planned paths as estimated by the GP model, as well as the path length (right axis) for different values of $\alpha$ (x-axis) used in the cost function during path planning.
As can be seen, the resulting path complexity drops as alpha is increased, and the path length increases with increase of $\alpha$. 
For this evaluation, we use $\alpha = 0.75$, which is at the inflection point.
For the results for $\alpha \in \{0,0.25,2,1000\}$, see appendix section~\ref{app:path_cost}, page~\pageref{app:path_cost}.

\begin{figure}[!t]
	\centering
	\includegraphics[width=.8\columnwidth]{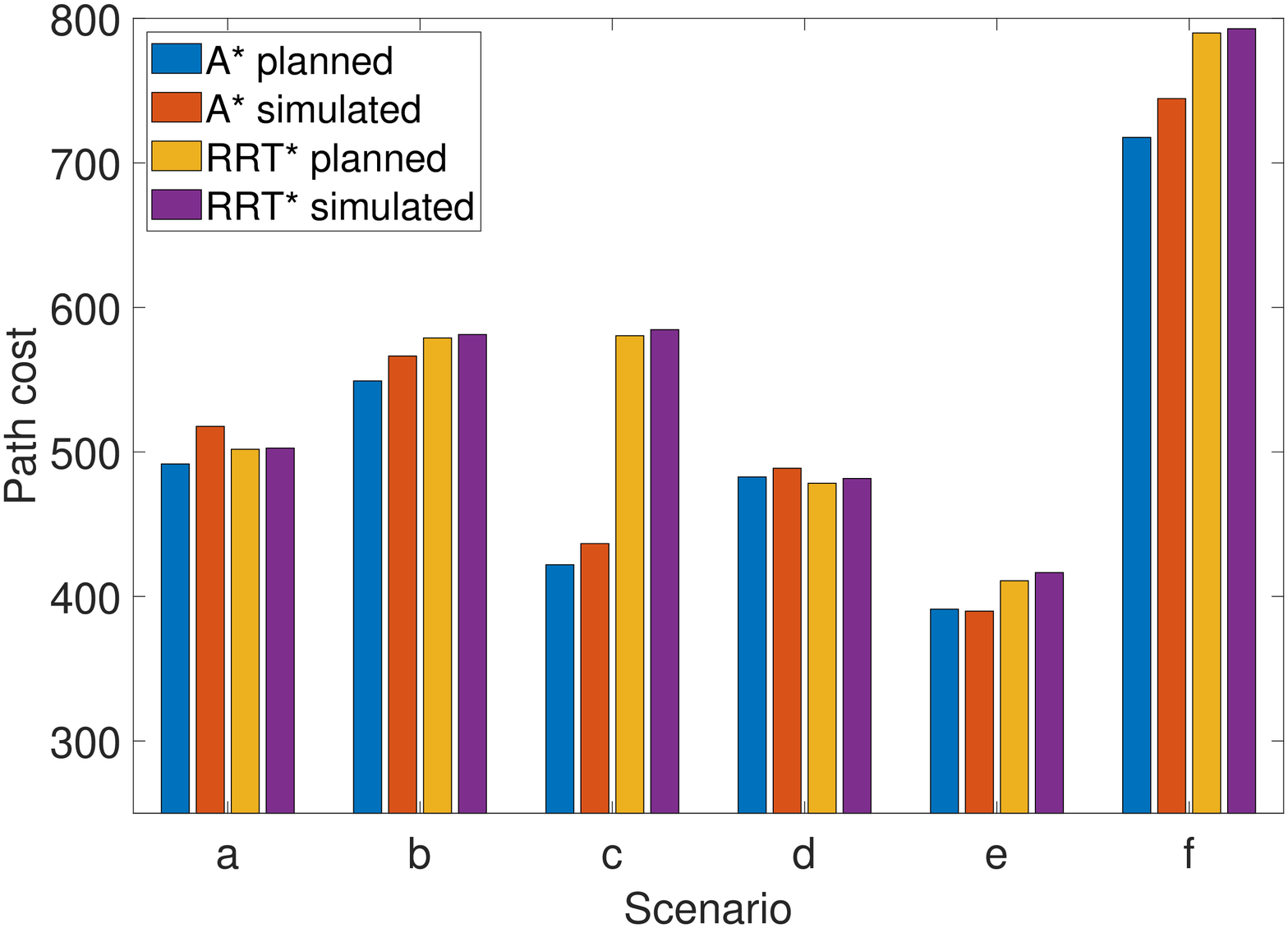}
	\caption{Path cost $g$, where $g = l + \alpha \chi$ for $\alpha=0.75$, for each path planner for the planned (blue, red) and simulated AUV (yellow, purple) paths. Paths are planned with $\alpha = 0.75$. Y-axis is cropped to start at 250.\vspace{-.1cm}}
	\label{fig:path_cost3}
\end{figure}

\begin{figure}[!t]
	\centering
	\includegraphics[width=.8\columnwidth]{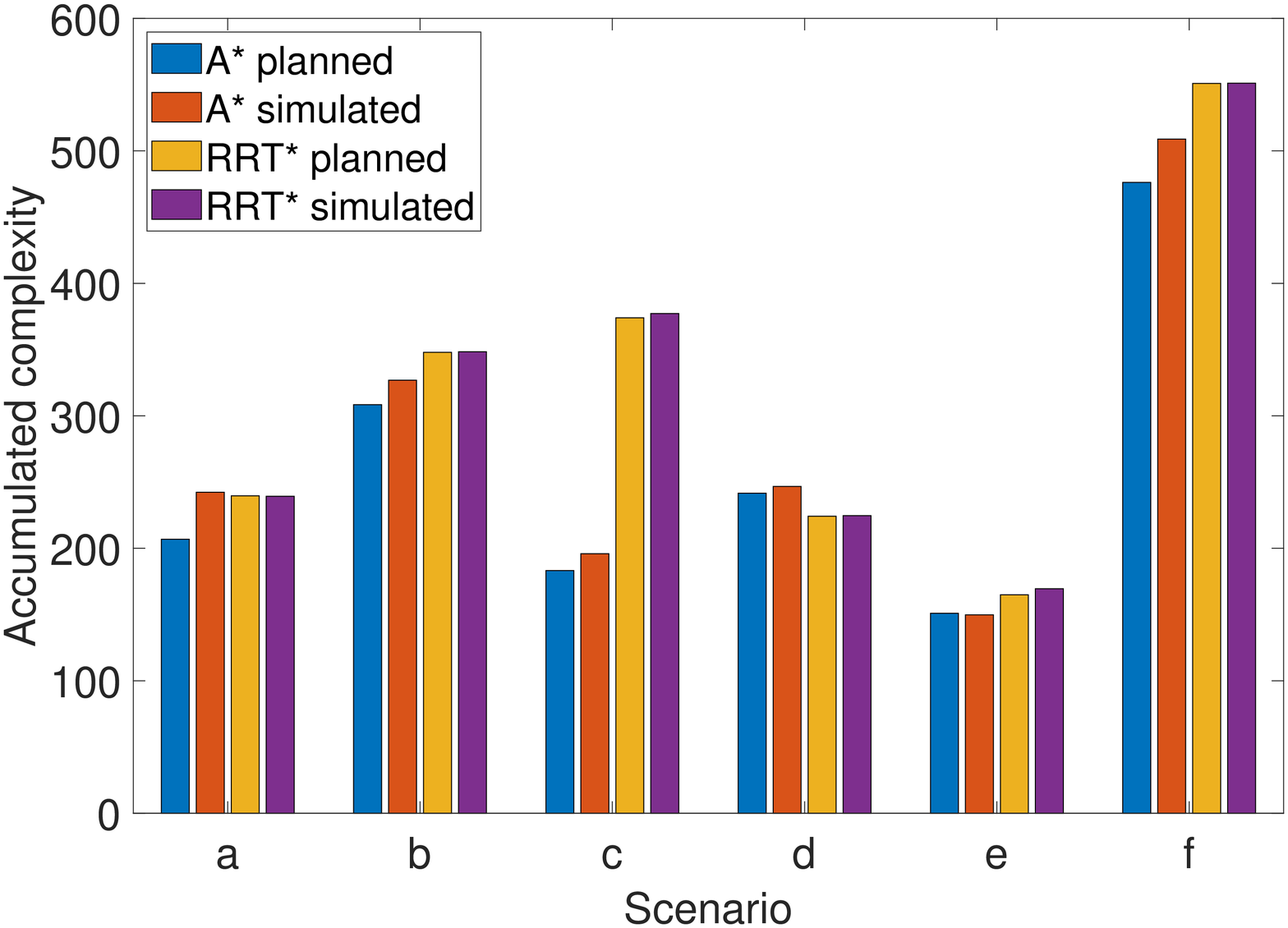}
	\caption{Accumulated complexity for each path planner for the planned (blue, red) and simulated AUV (yellow, purple) paths. Paths are planned with $\alpha = 0.75$.\vspace{-.1cm}}
	\label{fig:path_complexity3}
\end{figure}

\begin{figure}[!t]
	\centering
	\includegraphics[width=.8\columnwidth]{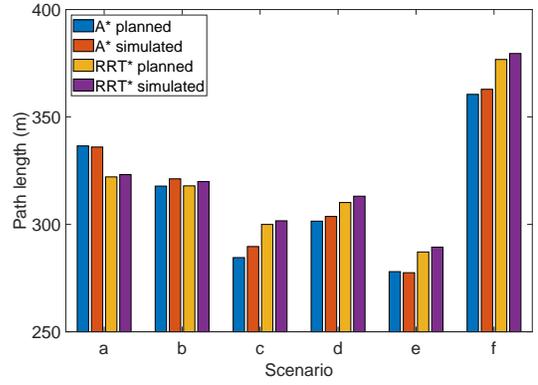}
	\caption{Path length for each path planner for the planned (blue, red) and simulated AUV (yellow, purple) paths. Paths are planned with $\alpha = 0.75$. Y-axis is cropped to start at 250.\vspace{-.1cm}}
	\label{fig:path_length3}
\end{figure}

Figure~\ref{fig:path_cost3} shows the path cost for all six scenarios and for each planner, for both the planned path and the path executed by the dynamic model simulation, for $\alpha = 0.75$.
The path cost incorporates both the path length, shown in Figure~\ref{fig:path_length3}, and the accumulated complexity as estimated by the GP model, shown in Figure~\ref{fig:path_complexity3}. 
We see that performance varies with each scenario, and overall the path planners produce similar results.
For scenario c), A* path planning clearly outperforms RRT* path planning under this cost function.
This is mostly due to the increase in path length for RRT* paths in that scenario, as further investigated below.
Overall, we see that the difference between planned and executed paths is smaller for RRT*. This is due to it better considering vehicle dynamics by Dubins curve integration.

In terms of path length, for scenarios c-f, RRT* produces longer paths than A*. 
This may also be due to the inclusion of vehicle dynamics via Dubins curves in the RRT path planning. 
If we compare the estimated accumulated complexity along the paths, Figure~\ref{fig:path_complexity3}, we notice that performance is varied across scenarios.
Most notable here is that the accumulated complexity increases when comparing simulated to planned trajectories.
This is the case in particular for A*, which may be due to paths that are planned close to higher complexity areas, and where the restrictions imposed by vehicle dynamics force the vehicle to go through these areas more than planned.
Because RRT* better incorporates vehicle dynamics, difference in estimated complexity values between planned and executed are smaller.


\begin{figure}[!thb]
	\centering
	\includegraphics[width=\columnwidth]{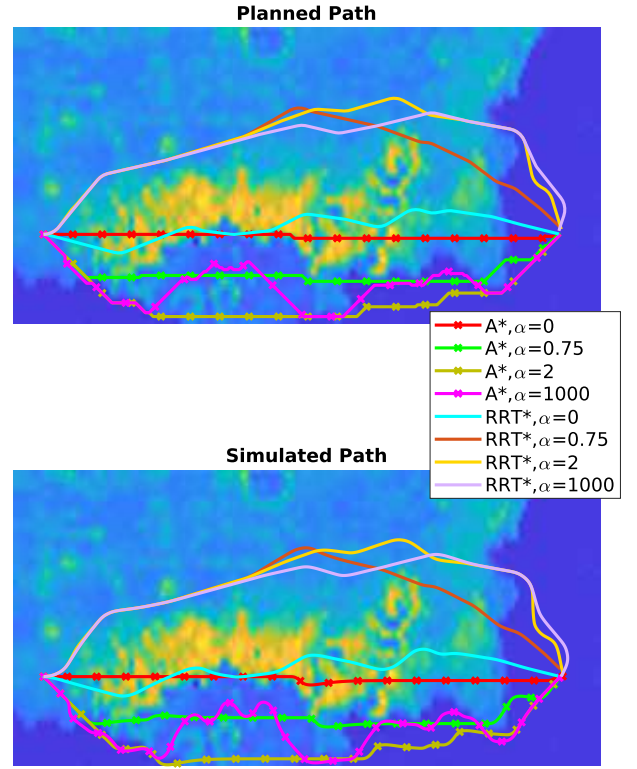} 
	\caption{Planned and simulated paths for scenario c), comparing the path planning methods for four values of $\alpha$.}
	\label{fig:all_paths_c}
\end{figure}

To compare the effect of the parameter $\alpha$, we also show the planned paths for scenario c) for  four values of alpha, $alpha = {0, 0.75, 2, 1000}$, in Figure~\ref{fig:all_paths_c}. 
As $\alpha$ increases, the path complexity becomes more important to the path cost than the path length, and trajectories diverge more from the shortest path ($\alpha = 0$) to seek out areas of low complexity.
It would be up to an end-user or vehicle operator to decide what risks are acceptable, and how much time can be spent traversing a route, to determine desirable trajectories.

\section{Discussion \& Future Work}
Our experiments show that for modeling seabed complexity data using GP regression, the additive Mat\'ern kernel is most suitable.
The results also show that the individual Mat\'ern kernels outperform the Squared Exponential (SE) kernel in some scenarios, and that the neural network kernel performs worst for most scenarios.
These findings suggest that the infinite differentiability of the SE kernel may be problematic when trying to model physical phenomena, such as complexity in sonar images. 
Stein has noted that the infinite differentiability of the SE kernel "would normally be considered unrealistic for a physical process"~\cite{Stein1999} and suggests the Mat\'ern kernel for use in geospatial statistics.

In terms of path planning, the performance between RRT* with kinematic constraints and A* was scenario dependent.
This paper focused on the results for $\alpha=0.75$, but results for all values are shown in the Appendix.
The value of $\alpha=0.75$ provides a reasonable trade-off between path length and path complexity.
One clear result from the simulations was that the accumulated complexity changed between the planned and simulated paths, especially for A*. 
In future work, other planners which take into account vehicle dynamics more closely may be explored, beyond RRT* with Dubins curves.
One promising approach is the RRT* with kino-dynamic constraints as presented by Webb and van der Berg~\cite{Webb2013}. 
However this approach will be more computationally demanding.

In this work, we have created a GP model using data obtained from lawnmower surveys.
Part of the reason to use GP regression is to enable on-line path planning.
In future work, we plan to investigate how well safe paths can be found through environments if a full model of the environment is not yet available, and has to be constructed on the fly.

\section{Conclusion}
This work proposed the use of GP regression to model seabed complexity data, and explore different kernels to best model this data.
Furthermore, we looked at path planning methods that can use this model to find safe paths through hazardous regions. 
Our work shows that the commonly used Squared Exponential kernel performs worse than the proposed Mat\'ern Additive kernel for prediction of seabed complexity data. 
We compared five kernels on six scenarios, and show that the Mat\'ern Additive kernel outperforms the other kernels. 
We therefore recommend the use of the additive Mat\'ern kernel for modeling seabed complexity or bathymetry data.
Furthermore, we evaluated two path planners, RRT* with kinematic constraints and A*. Both path planners incorporate the estimates on seabed complexity from the GP model. We ran a vehicle dynamics simulator to compare between planned paths and paths as they might be executed by a marine vessel.
Overall, the path planner performance is similar. RRT* creates longer paths for four of the scenarios, which is related to it using the Dubins curve to model vehicle dynamics. 
At the same time, this integration of vehicle dynamics into the path planning leads to better performance with the vehicle dynamics simulator.
For most scenarios, the path length and complexity increase a little after we run the vehicle dynamics model, which shows that the path planning methods could be improved further by further incorporating vehicle dynamics models, especially for the A* method.


\section*{Acknowledgments}
The authors thank Marc Geilhufe (Norwegian Defence Research Establishment, FFI) for providing complexity data from the HUGIN AUV.

\bibliographystyle{IEEEtran}
\bibliography{references}

\begin{thebibliography}{10}
\providecommand{\url}[1]{#1}
\csname url@rmstyle\endcsname
\providecommand{\newblock}{\relax}
\providecommand{\bibinfo}[2]{#2}
\providecommand\BIBentrySTDinterwordspacing{\spaceskip=0pt\relax}
\providecommand\BIBentryALTinterwordstretchfactor{4}
\providecommand\BIBentryALTinterwordspacing{\spaceskip=\fontdimen2\font plus
\BIBentryALTinterwordstretchfactor\fontdimen3\font minus
  \fontdimen4\font\relax}
\providecommand\BIBforeignlanguage[2]{{%
\expandafter\ifx\csname l@#1\endcsname\relax
\typeout{** WARNING: IEEEtran.bst: No hyphenation pattern has been}%
\typeout{** loaded for the language `#1'. Using the pattern for}%
\typeout{** the default language instead.}%
\else
\language=\csname l@#1\endcsname
\fi
#2}}

\bibitem{Geilhufe2016}
M.~Geilhufe, W.~A. Connors, and {\O}.~Midtgaard, ``{Through-the-Sensor
  Performance Evaluation for Modern Mine Hunting Operations},'' in \emph{UDT
  Europe}, 2016.

\bibitem{Hagen2003}
P.~E. Hagen, N.~Storkersen, K.~Vestgard, and P.~Kartvedt, ``{The {HUGIN} 1000
  Autonomous Underwater Vehicle for Military Applications},'' in \emph{Oceans
  2003. Proceedings}, vol.~2.\hskip 1em plus 0.5em minus 0.4em\relax IEEE,
  2003, pp. 1141--1145.

\bibitem{Chang2006}
K.-T. Chang, \emph{{Introduction to Geographic Information Systems}}.\hskip 1em
  plus 0.5em minus 0.4em\relax {McGraw-Hill}, 2006.

\bibitem{Tobler1970}
W.~R. Tobler, ``{A Computer Movie Simulating Urban Growth in the Detroit
  Region},'' \emph{Economic Geography}, vol.~46, pp. 234--240, 1970.

\bibitem{Krause2008}
A.~Krause, A.~Singh, and C.~Guestrin, ``{Near-Optimal Sensor Placements in
  Gaussian Processes-Theory, Efficient Algorithms and Empirical Studies},''
  \emph{The Journal of Machine Learning Research}, vol.~9~, no. May, pp.
  235--284, 2008.

\bibitem{Low2008}
K.~H. Low, J.~M. Dolan, and P.~Khosla, ``{Adaptive multi-robot wide-area
  exploration and mapping},'' in \emph{Proceedings of the 7th international
  joint conference on Autonomous agents and Multiagent systems}.\hskip 1em plus
  0.5em minus 0.4em\relax International Foundation for Autonomous Agents and
  Multiagent Systems, 2008, pp. 23--30.

\bibitem{Yang2013}
K.~Yang, S.~K. Gan, and S.~Sukkarieh, ``{A Gaussian process-based RRT planner
  for the exploration of an unknown and cluttered environment with a UAV},''
  \emph{Advanced Robotics}, vol.~27, no.~6, pp. 431--443, 2013.

\bibitem{Hollinger2013}
G.~A. Hollinger, A.~A. Pereira, and G.~S. Sukhatme, ``{Learning uncertainty
  models for reliable operation of autonomous Underwater Vehicles},'' in
  \emph{2013 IEEE International Conference on Robotics and Automation (ICRA)},
  May 2013, pp. 5593--5599.

\bibitem{Vasudevan2009}
S.~Vasudevan, F.~Ramos, E.~Nettleton, H.~Durrant-Whyte, and A.~Blair,
  ``{Gaussian Process modeling of large scale terrain},'' in \emph{2009 IEEE
  International Conference on Robotics and Automation (ICRA)}, May 2009, pp.
  1047--1053.

\bibitem{Hollinger2014}
G.~A. Hollinger and G.~S. Sukhatme, ``{Sampling-based robotic information
  gathering algorithms},'' \emph{IJRR}, vol.~33, pp. 1271--1287, 2014.

\bibitem{Galceran2013}
E.~Galceran and M.~Carreras, ``{A survey on coverage path planning for
  robotics},'' \emph{Robotics and Autonomous Systems}, vol.~61, no.~12, pp.
  1258--1276, Dec. 2013.

\bibitem{Paull2014}
L.~Paull, M.~Seto, and H.~Li, ``{Area coverage planning that accounts for pose
  uncertainty with an {AUV} seabed surveying application},'' in \emph{2014 IEEE
  International Conference on Robotics and Automation (ICRA)}, 2014, pp.
  6592--6599.

\bibitem{Krogstad2014}
T.~R. Krogstad and M.~S. Wiig, ``{Autonomous Survey and Identification Planning
  for {AUV} {MCM} Operations},'' in \emph{UDT Europe}, 2014.

\bibitem{Williams2010}
D.~P. Williams, ``{On optimal AUV track-spacing for underwater mine
  detection},'' in \emph{2010 IEEE International Conference on Robotics and
  Automation (ICRA)}.\hskip 1em plus 0.5em minus 0.4em\relax IEEE, 2010, pp.
  4755--4762.

\bibitem{Bekker2006}
J.~Bekker and J.~Schmid, ``{Planning the safe transit of a ship through a
  mapped minefield},'' \emph{ORiON}, vol.~22, no.~1, 2006.

\bibitem{Rasmussen2006}
C.~Rasmussen and C.~Williams, \emph{{Gaussian Processes for Machine
  Learning.}}\hskip 1em plus 0.5em minus 0.4em\relax {MIT press}, 2006.

\bibitem{Stein1999}
M.~L. Stein, \emph{{Interpolation of Spatial Data: Some Theory for
  Kriging}}.\hskip 1em plus 0.5em minus 0.4em\relax Springer Science+Business
  Media New York, 1999.

\bibitem{Choset2000}
H.~Choset, ``{Coverage of Known Spaces: The Boustrophedon Cellular
  Decomposition},'' \emph{Autonomous Robots}, vol.~9, no.~3, pp. 247--253, Dec
  2000.

\bibitem{gpml}
\BIBentryALTinterwordspacing
C.~Rasmussen, H.~Nickish, and C.~Williams, ``{GPML Matlab Code version 4.1},''
  [Online], last retrieved November, 2017. [Online]. Available:
  \url{http://www.gaussianprocess.org/gpml/code/matlab/doc/}
\BIBentrySTDinterwordspacing

\bibitem{Hart1968}
P.~E. Hart, N.~J. Nilsson, and B.~Raphael, ``{A Formal Basis for the Heuristic
  Determination of Minimum Cost Paths},'' \emph{IEEE transactions on Systems
  Science and Cybernetics}, vol.~4, no.~2, pp. 100--107, 1968.

\bibitem{Lavalle1998}
S.~M. LaValle, ``{Rapidly-exploring random trees: A new tool for path
  planning},'' Computer Science Department, Iowa State University, Report TR
  98-11, 1998.

\bibitem{Karaman2011}
S.~Karaman and E.~Frazzoli, ``{Sampling-based algorithms for optimal motion
  planning},'' \emph{The international journal of robotics research}, vol.~30,
  no.~7, pp. 846--894, 2011.

\bibitem{Karaman2011a}
S.~Karaman, M.~R. Walter, A.~Perez, E.~Frazzoli, and S.~Teller, ``{Anytime
  motion planning using the {RRT}},'' in \emph{2011 IEEE International
  Conference on Robotics and Automation (ICRA)}.\hskip 1em plus 0.5em minus
  0.4em\relax IEEE, 2011, pp. 1478--1483.

\bibitem{Dubins1957}
L.~E. Dubins, ``{On curves of minimal length with a constraint on average
  curvature, and with prescribed initial and terminal positions and
  tangents},'' \emph{American Journal of Mathematics}, vol.~79, no.~3, pp.
  497--516, 1957.

\bibitem{Fossen2011}
T.~I. Fossen, \emph{{Handbook of Marine Craft, Hydrodynamics and Motion
  Control}}.\hskip 1em plus 0.5em minus 0.4em\relax John Wiley \& Sons, 2011.

\bibitem{Moe2014}
S.~Moe, W.~Caharija, K.~Y. Pettersen, and I.~Schjolberg, ``{Path Following of
  Underactuated Marine Surface Vessels in the Presence of Unknown Ocean
  Currents},'' in \emph{American Control Conference (ACC), 2014}.\hskip 1em
  plus 0.5em minus 0.4em\relax IEEE, 2014, pp. 3856--3861.

\bibitem{Webb2013}
D.~J. Webb and J.~van~den Berg, ``{Kinodynamic {RRT*}: Asymptotically optimal
  motion planning for robots with linear dynamics},'' in \emph{2013 IEEE
  International Conference on Robotics and Automation (ICRA)}.\hskip 1em plus
  0.5em minus 0.4em\relax IEEE, 2013, pp. 5054--5061.

\end{thebibliography}


\cleardoublepage

\section*{Appendix}

In this appendix we list all additional figures for the results from the kernel comparison, and for the results from the path planning simulations.

~\\

\section{Kernel comparison}\label{app:kernel}

\begin{figure}[!h]
  \centering
  \includegraphics[width=\columnwidth]{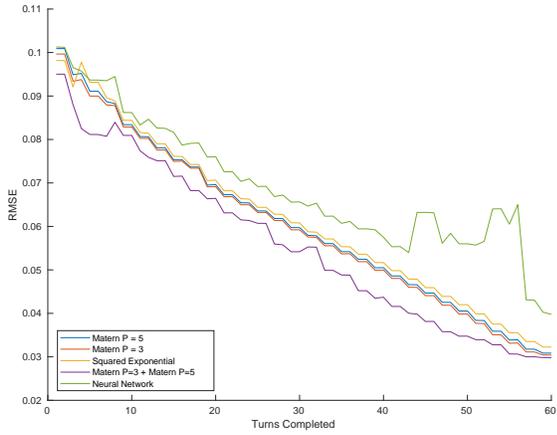} 
  \caption{RMSE curve for all kernels on scenario a) (Figure~\ref{fig:scenarios}).}
  \label{fig:gp_rmse_a}
\end{figure}

\begin{figure}[!h]
  \centering
  \includegraphics[width=\columnwidth]{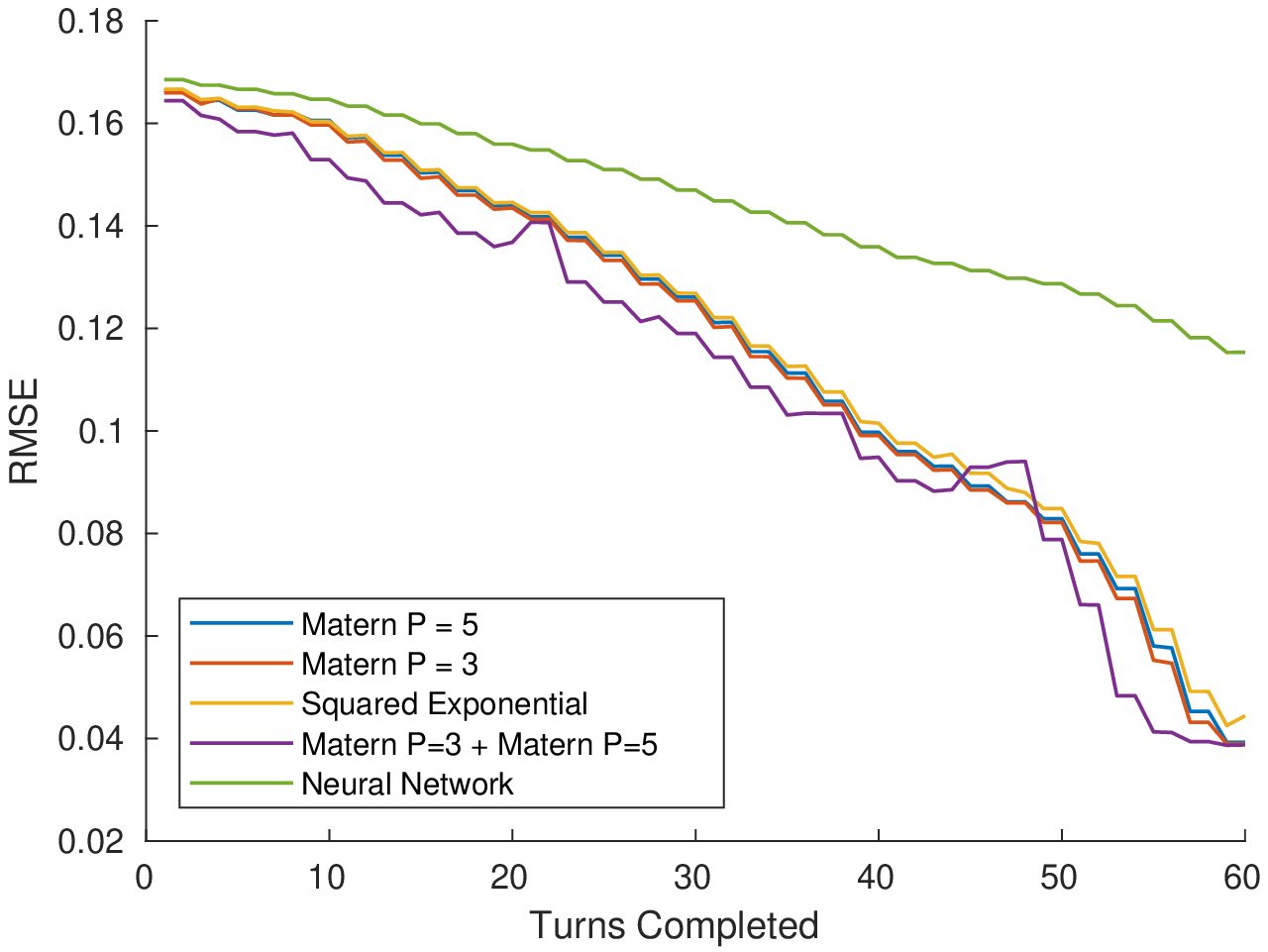} 
  \caption{RMSE curve for all kernels on scenario c) (Figure~\ref{fig:scenarios}).}
  \label{fig:gp_rmse_c}
\end{figure}

\begin{figure}[!h]
  \centering
  \includegraphics[width=\columnwidth]{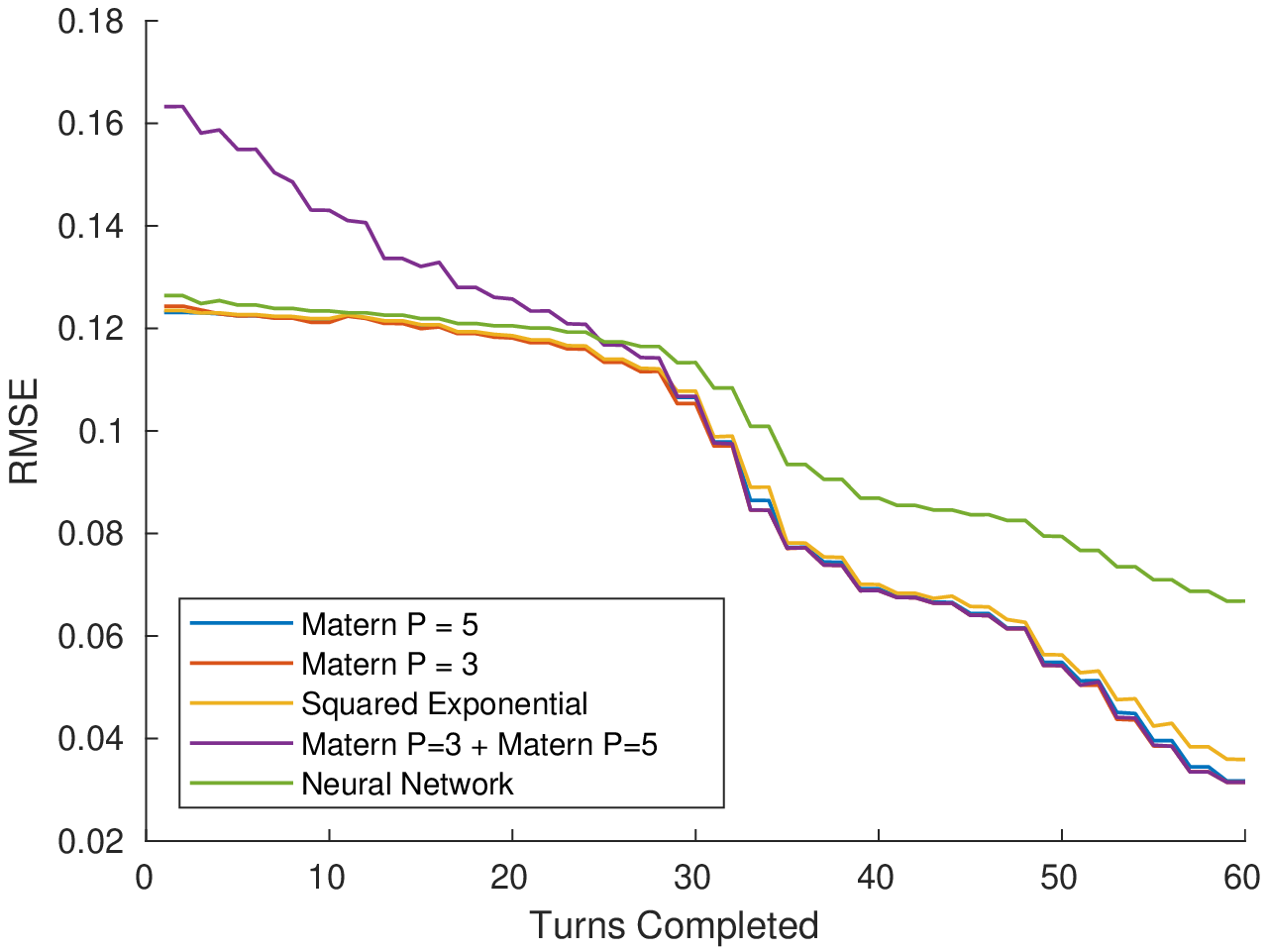} 
  \caption{RMSE curve for all kernels on scenario d) (Figure~\ref{fig:scenarios}).}
  \label{fig:gp_rmse_d}
\end{figure}

\begin{figure}[!h]
  \centering
  \includegraphics[width=\columnwidth]{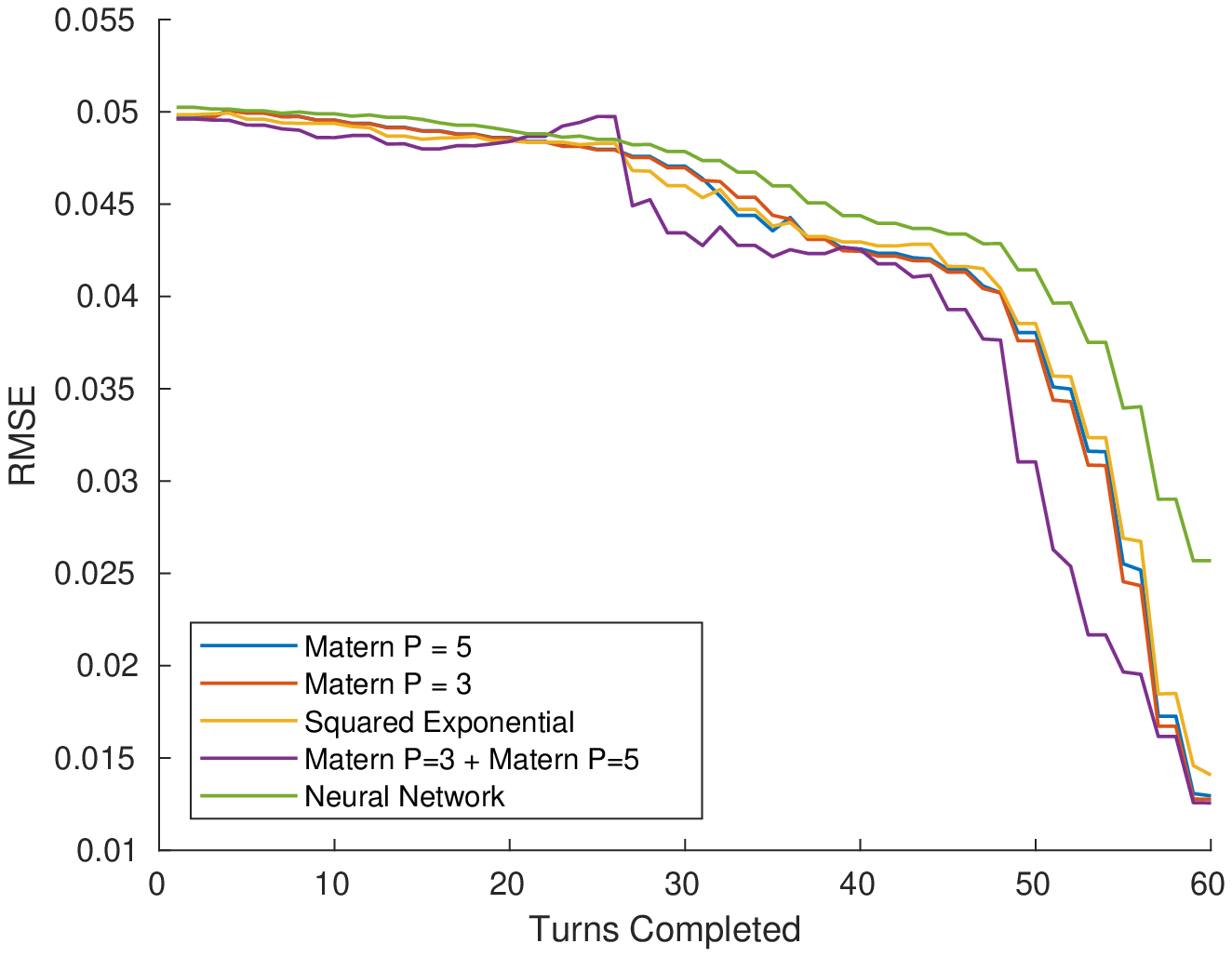} 
  \caption{RMSE curve for all kernels on scenario e) (Figure~\ref{fig:scenarios}).}
  \label{fig:gp_rmse_e}
\end{figure}

\begin{figure}[!h]
  \centering
  \includegraphics[width=\columnwidth]{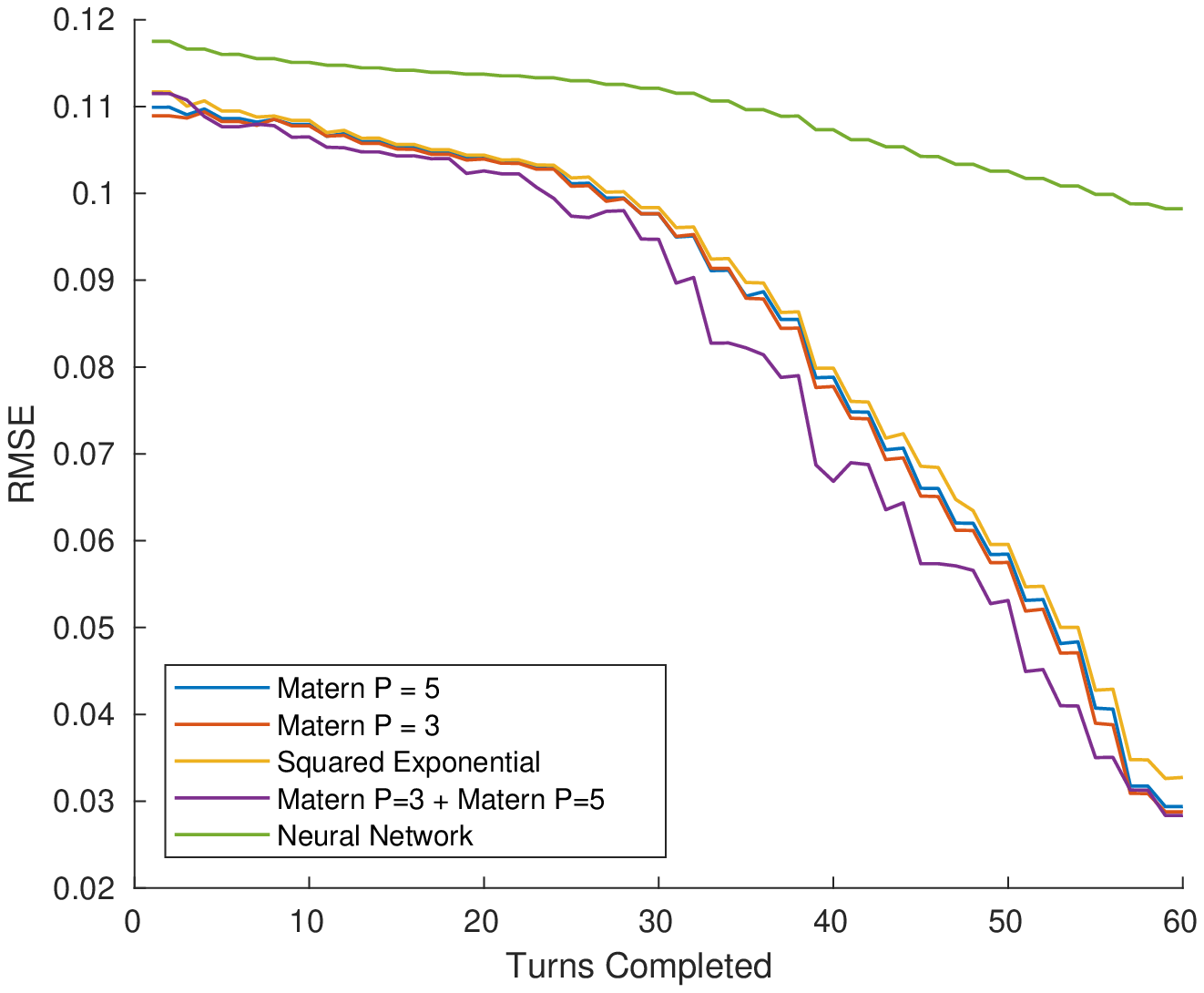} 
  \caption{RMSE curve for all kernels on scenario f) (Figure~\ref{fig:scenarios}).}
  \label{fig:gp_rmse_f}
\end{figure}

\cleardoublepage
\section{Path cost comparison}\label{app:path_cost}

{\bf Results for  $\alpha = 0$}
\begin{figure}[H]
	\centering
	\includegraphics[width=.85\columnwidth]{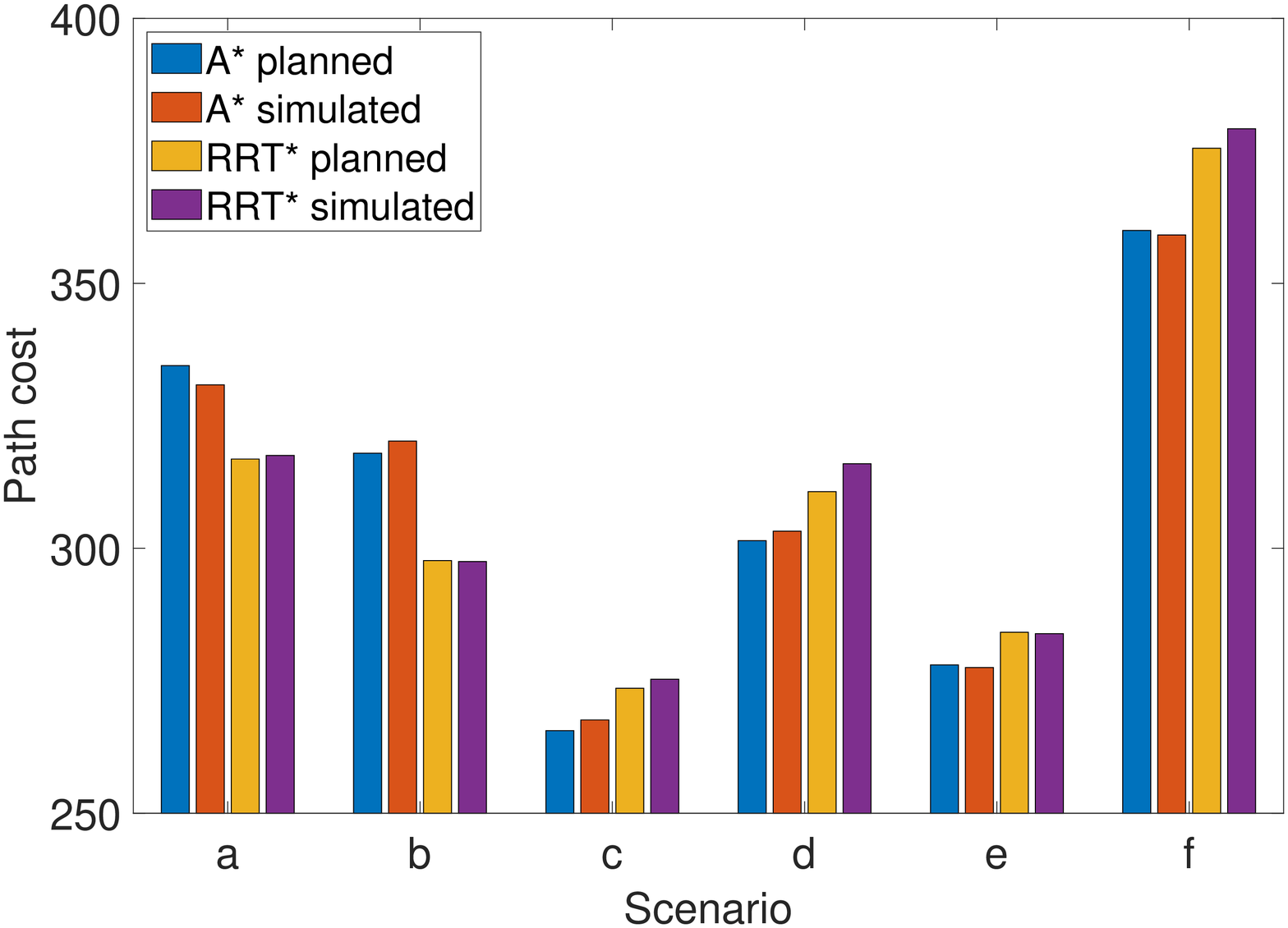}
	\caption{Path cost $g$, where $g = l + \alpha \chi$ for $\alpha=0$, for each path planner for the planned (blue, red) and simulated AUV (yellow, purple) paths. Paths are planned with $\alpha = 0$.}
	\label{fig:path_cost1}
\end{figure}
\begin{figure}[H]
	\centering
	\includegraphics[width=.85\columnwidth]{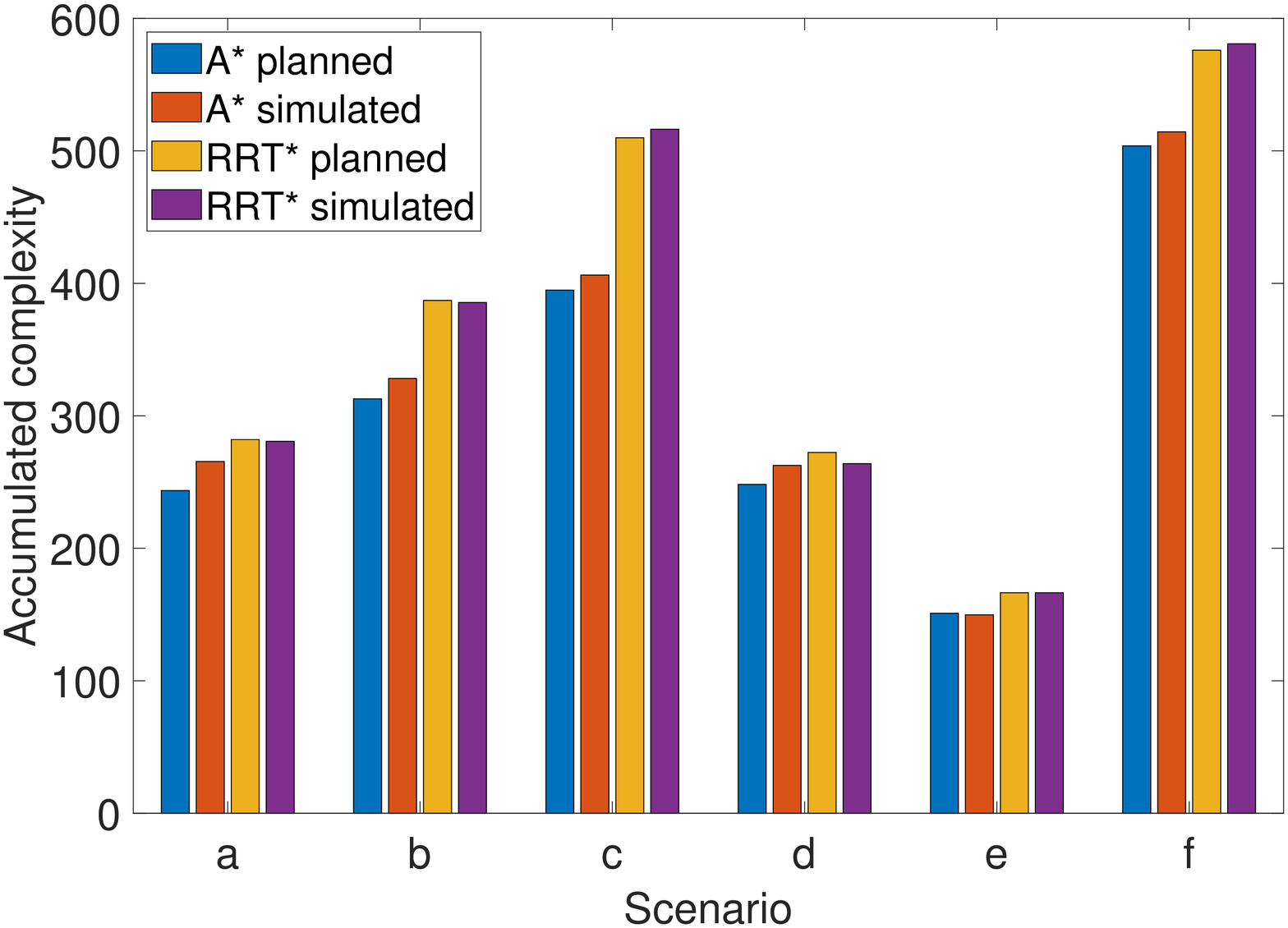}
	\caption{Accumulated complexity for each path planner for the planned (blue, red) and simulated AUV (yellow, purple) paths. Paths are planned with $\alpha = 0$.}
	\label{fig:path_complexity1}
\end{figure}
\begin{figure}[H]
	\centering
	\includegraphics[width=.85\columnwidth]{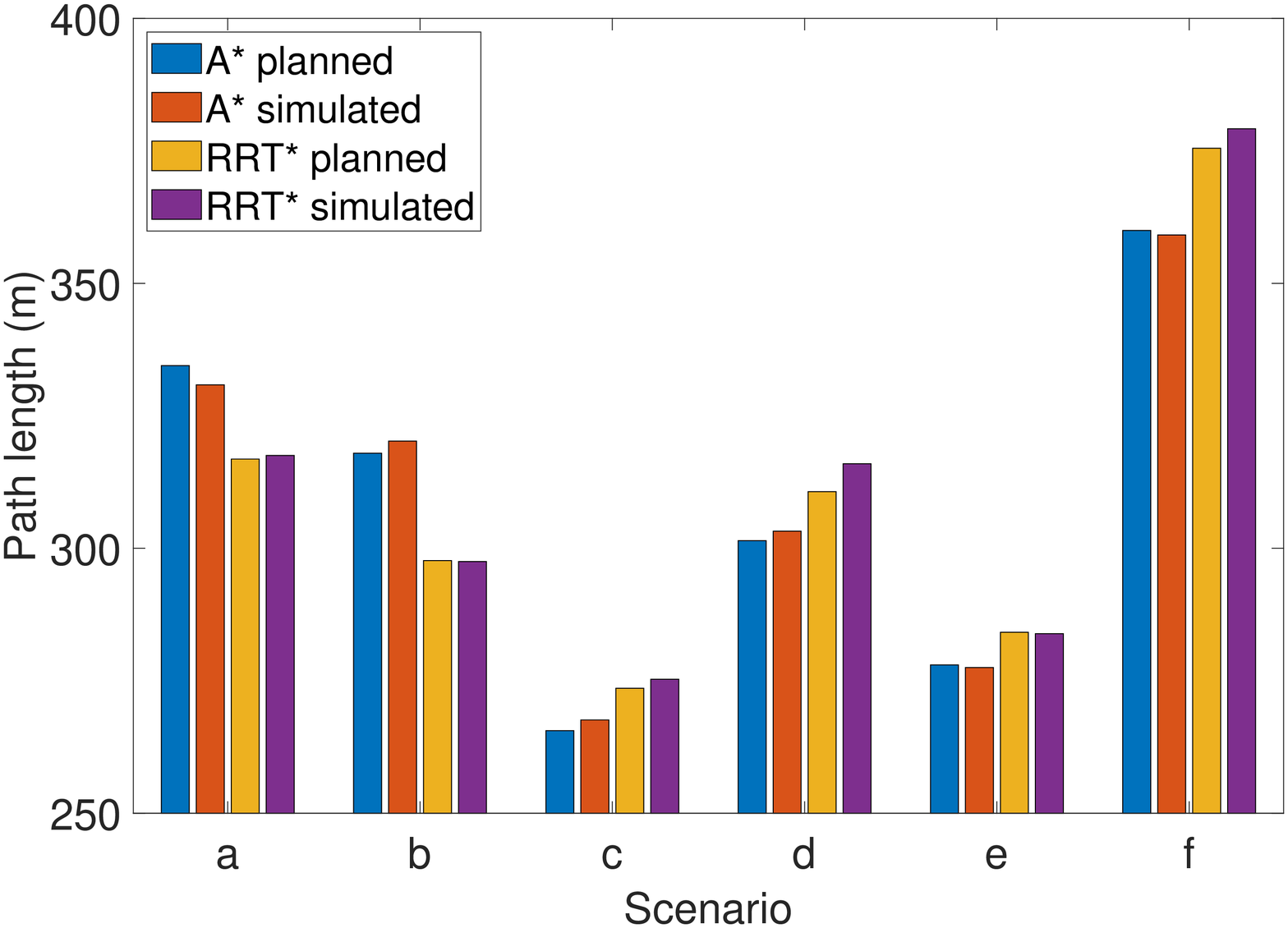}
	\caption{Path length comparison for each path planner for the planned (blue, red) and simulated AUV (yellow, purple) paths. Paths are planned with $\alpha = 0$.}
	\label{fig:path_length1}
\end{figure}

\newpage
\vspace{2cm}~~
~\\
{\bf Results for  $\alpha = 0.25$}
\begin{figure}[H]
	\centering
	\includegraphics[width=.85\columnwidth]{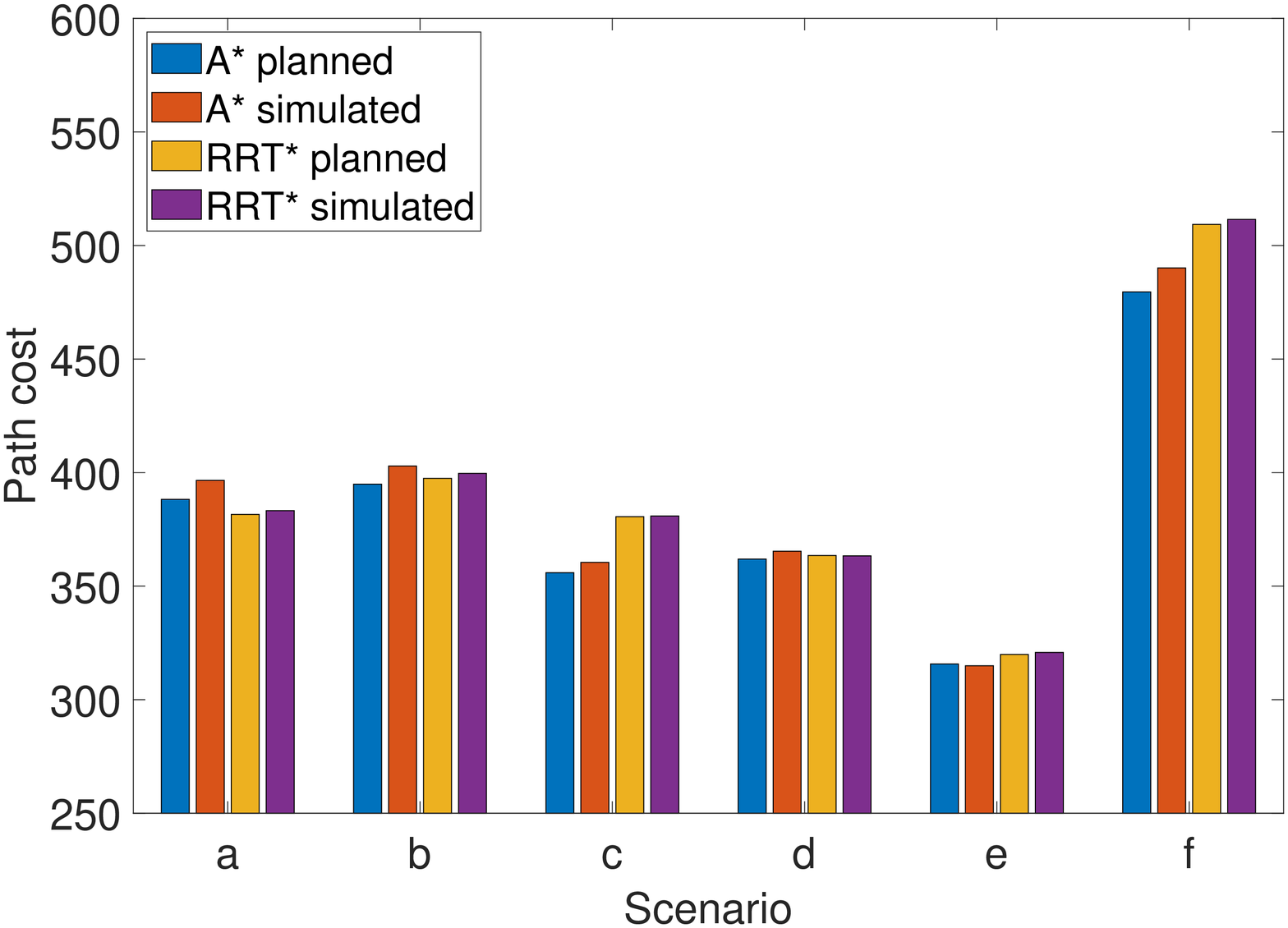}
	\caption{Path cost $g$, where $g = l + \alpha \chi$ for $\alpha=0.25$, for each path planner for the planned (blue, red) and simulated AUV (yellow, purple) paths. Paths are planned with $\alpha = 0.25$.}
	\label{fig:path_cost2}
\end{figure}
\begin{figure}[H]
	\centering
	\includegraphics[width=.85\columnwidth]{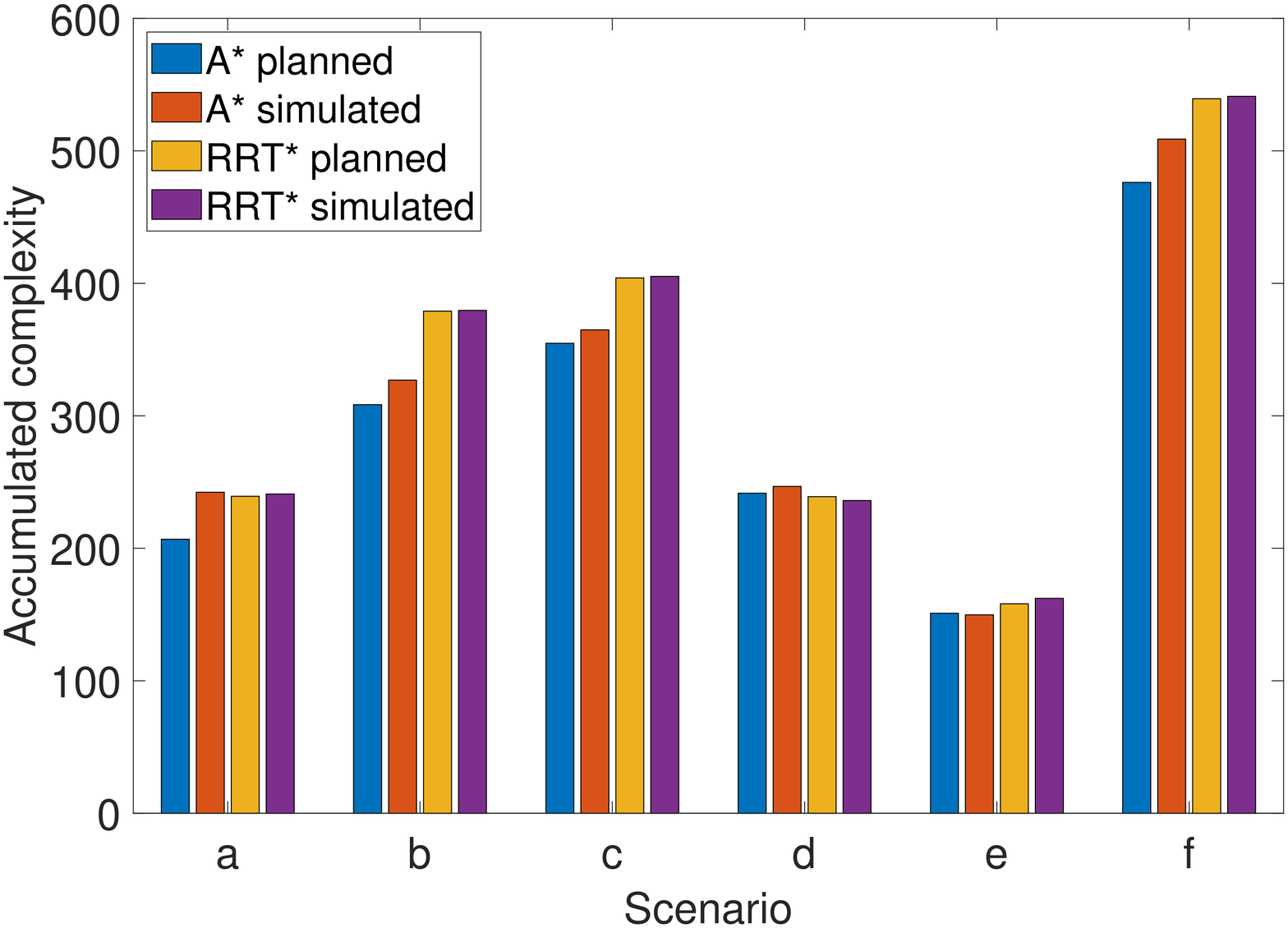} 
	\caption{Accumulated complexity for each path planner for the planned (blue, red) and simulated AUV (yellow, purple) paths. Paths are planned with $\alpha = 0.25$.}
	\label{fig:path_complexity2}
\end{figure}
\begin{figure}[H]
	\centering
	\includegraphics[width=.85\columnwidth]{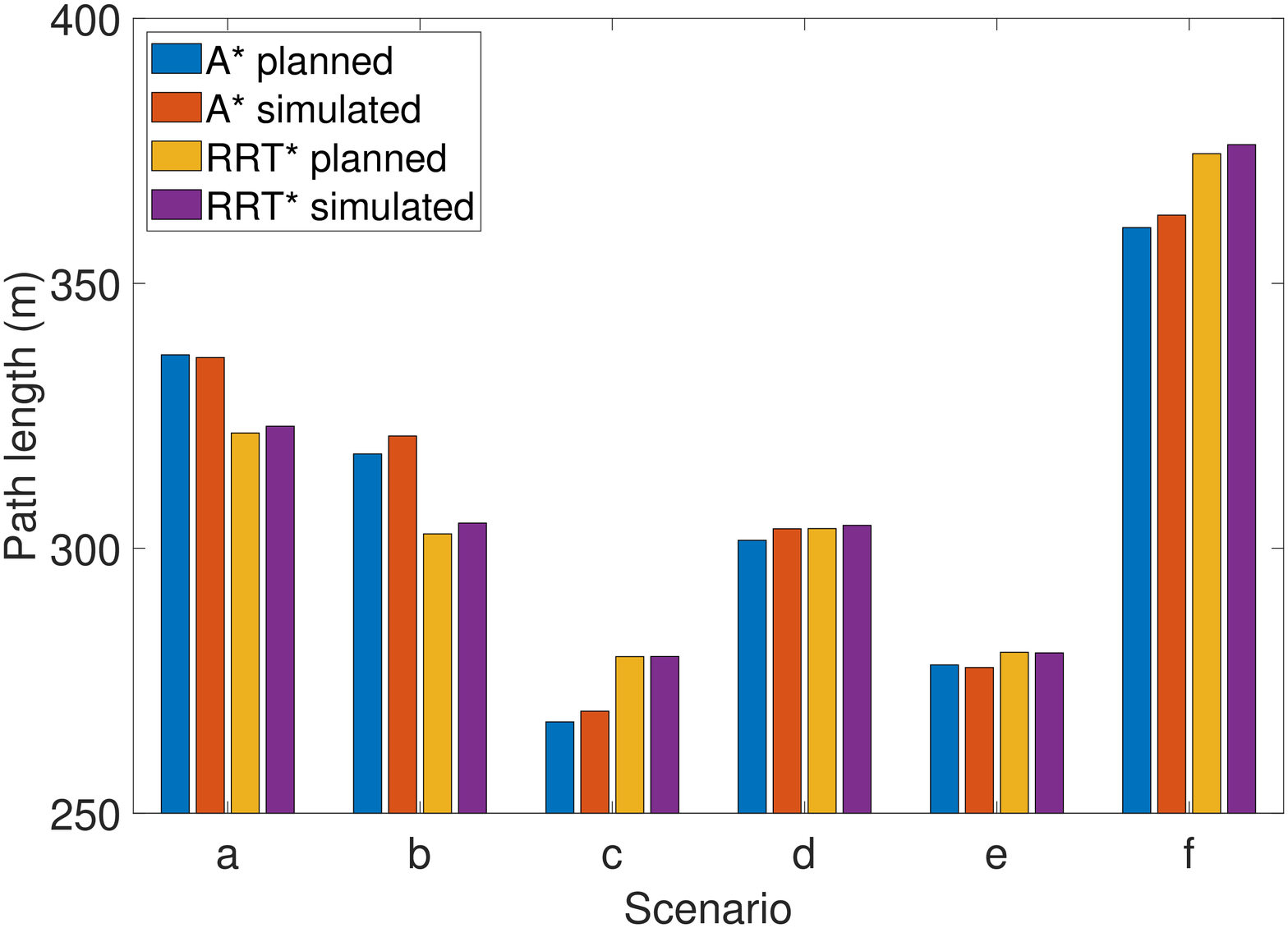} 
	\caption{Path length comparison for each path planner for the planned (blue, red) and simulated AUV (yellow, purple) paths. Paths are planned with $\alpha = 0.25$.}
	\label{fig:path_length2}
\end{figure}

\newpage
~~
~\\
{\bf Results for  $\alpha = 2$}\\
\begin{figure}[H]
	\centering
	\includegraphics[width=.85\columnwidth]{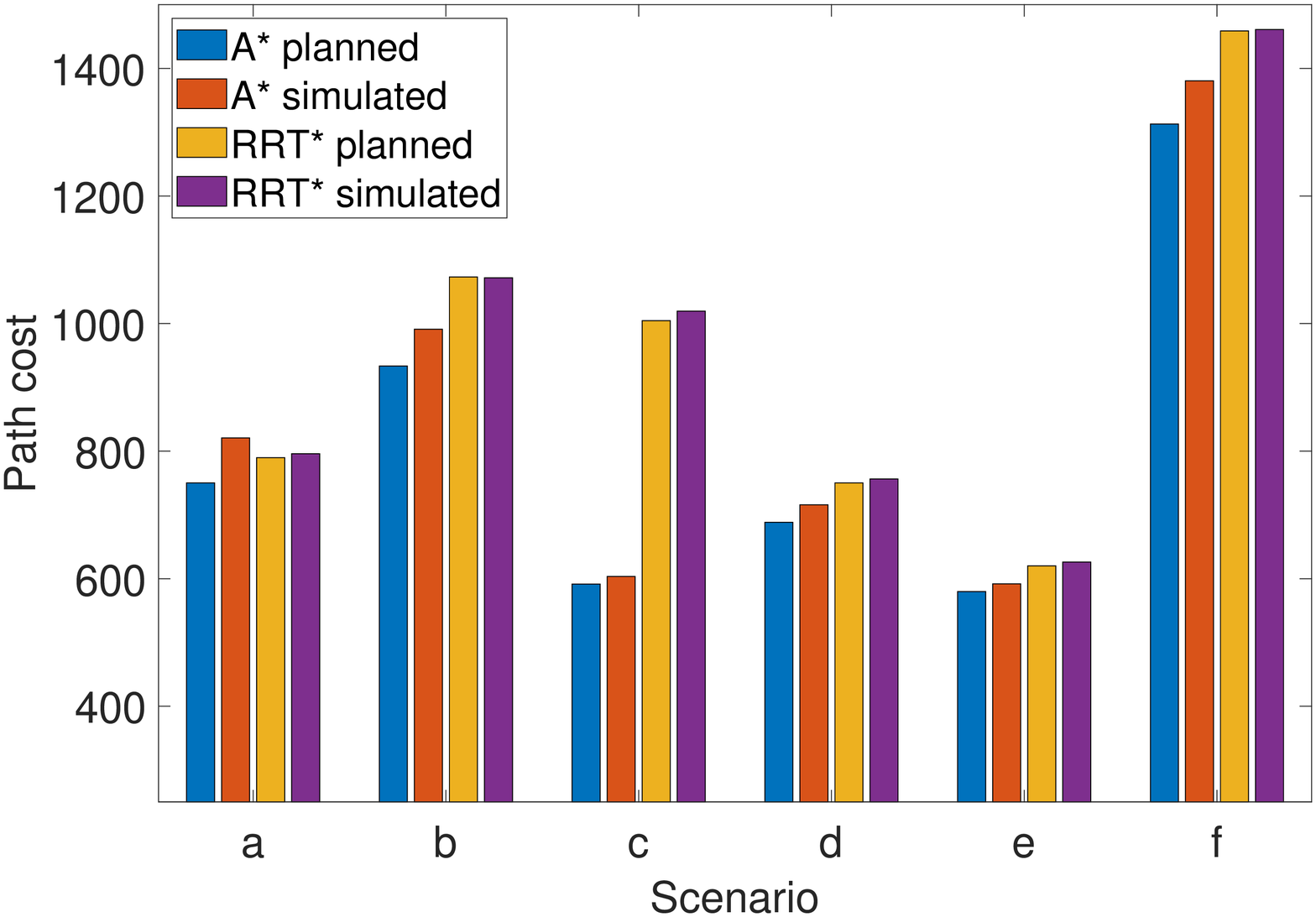}
	\caption{Path cost $g$, where $g = l + \alpha \chi$ for $\alpha=2$, for each path planner for the planned (blue, red) and simulated AUV (yellow, purple) paths. Paths are planned with $\alpha = 2$.}
	\label{fig:path_cost4}
\end{figure}
\begin{figure}[H]
	\centering
	\includegraphics[width=.85\columnwidth]{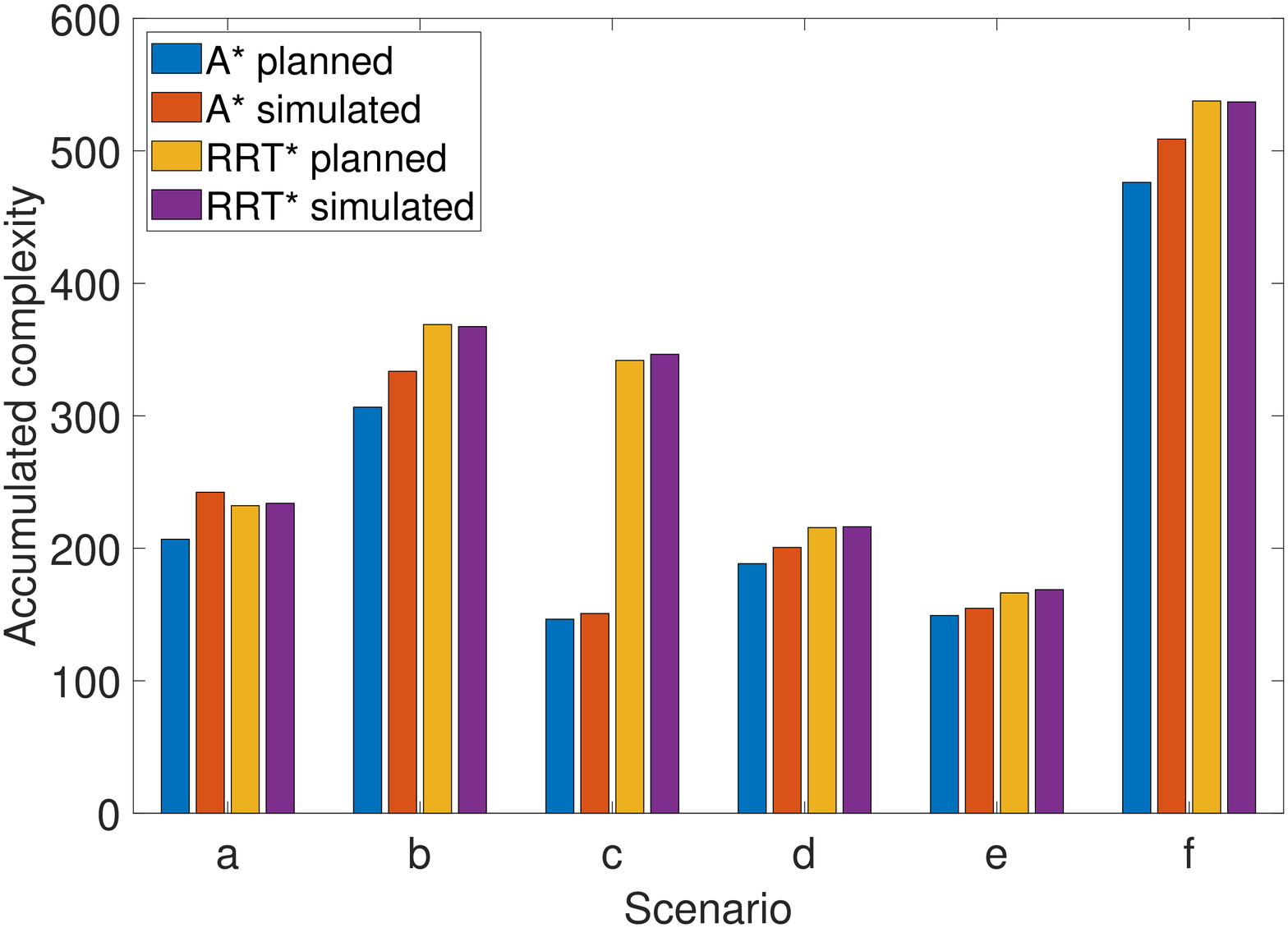}
	\caption{Accumulated complexity for each path planner for the planned (blue, red) and simulated AUV (yellow, purple) paths. Paths are planned with $\alpha = 2$.}
	\label{fig:path_complexity4}
\end{figure}
\begin{figure}[!h]
	\centering
	\includegraphics[width=.85\columnwidth]{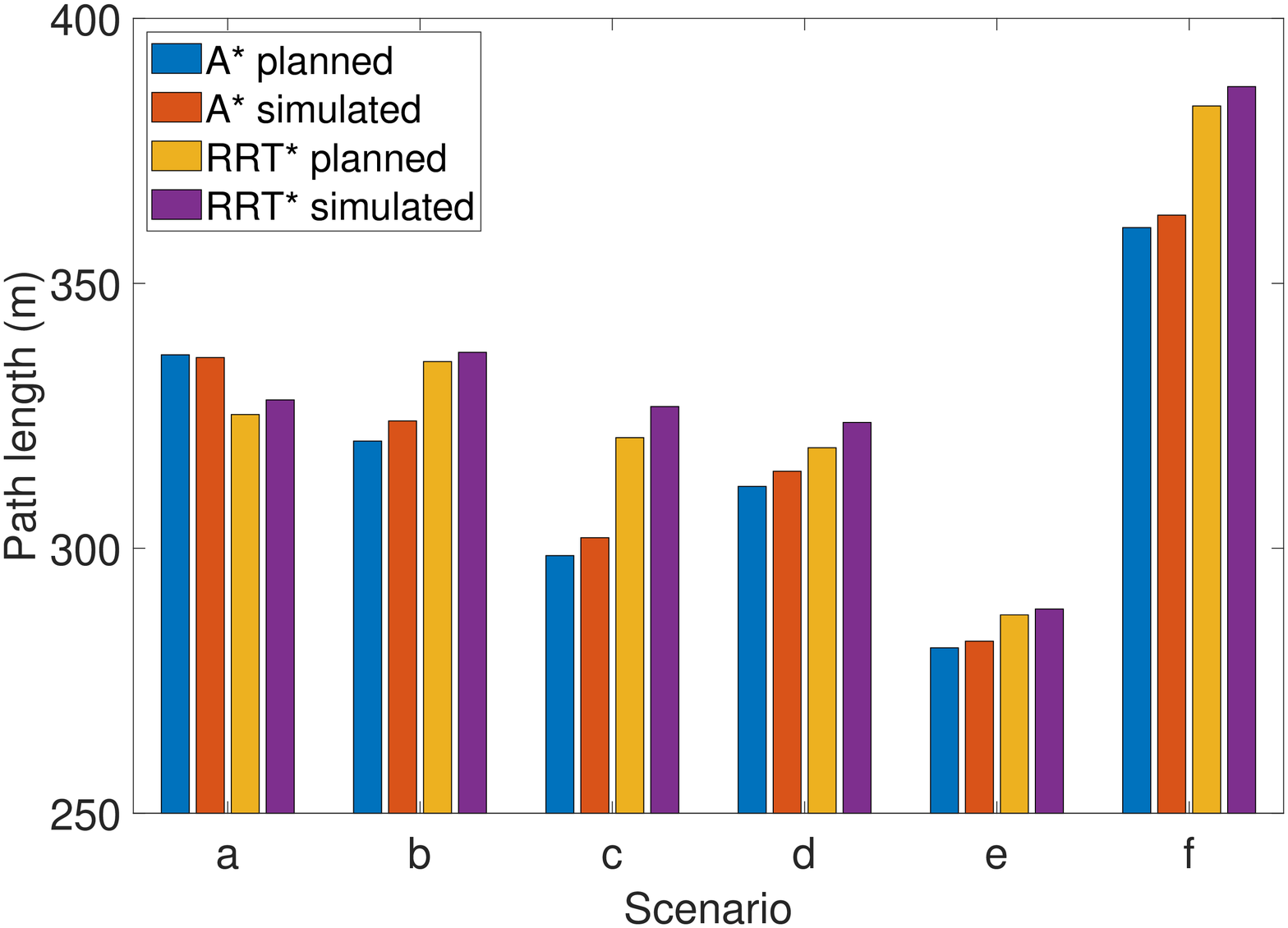}
	\caption{Path length comparison for each path planner for the planned (blue, red) and simulated AUV (yellow, purple) paths. Paths are planned with $\alpha = 2$.}
	\label{fig:path_length4}
\end{figure}

\newpage
~~
~\\
{\bf Results for  $\alpha = 1000$}
\begin{figure}[!h]
	\centering
	\includegraphics[width=.85\columnwidth]{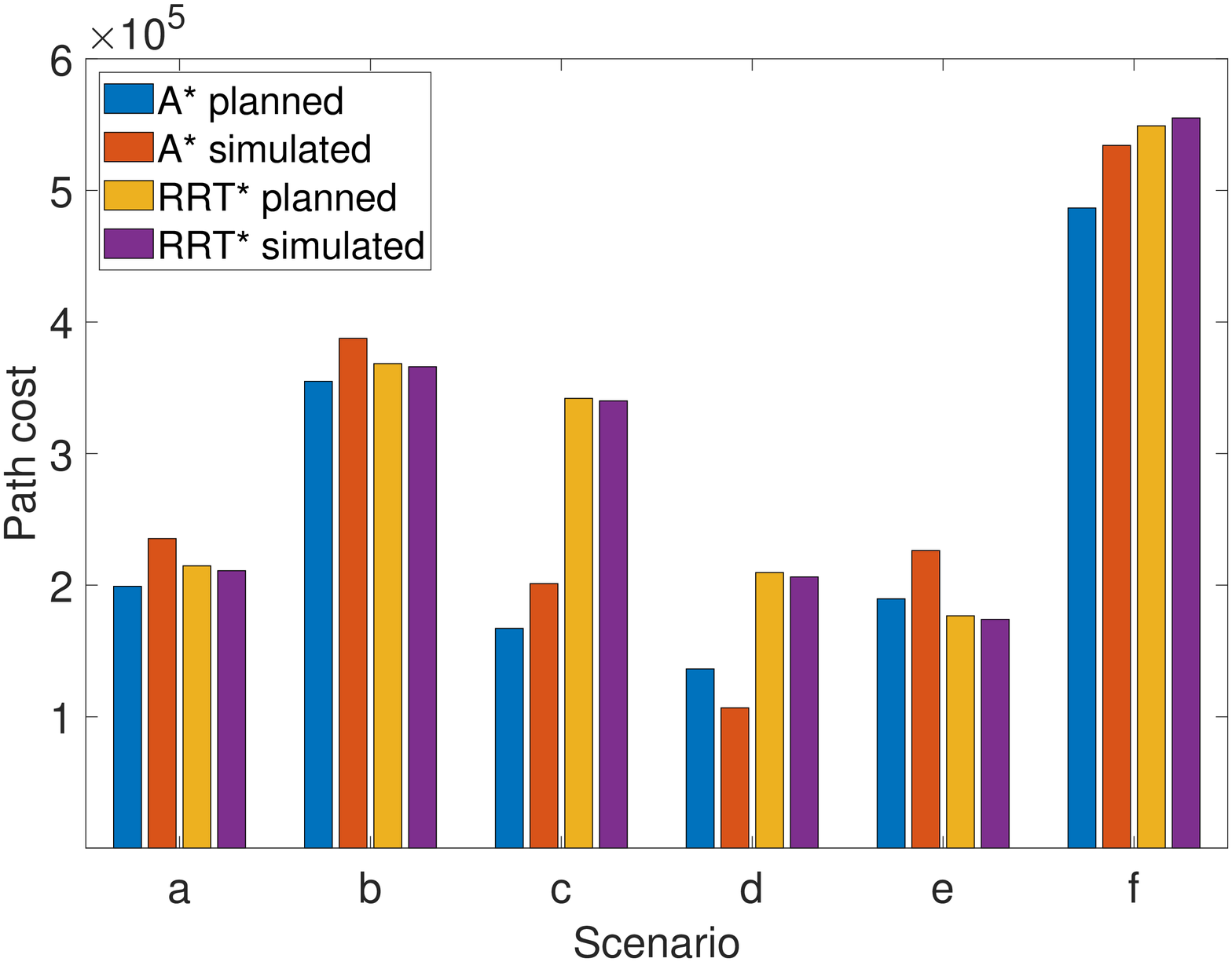}
	\caption{Path cost $g$, where $g = l + \alpha \chi$ for $\alpha=1000$, for each path planner for the planned (blue, red) and simulated AUV (yellow, purple) paths. Paths are planned with $\alpha = 1000$.}
	\label{fig:path_cost5}
\end{figure}
\begin{figure}[!h]
	\centering
	\includegraphics[width=.85\columnwidth]{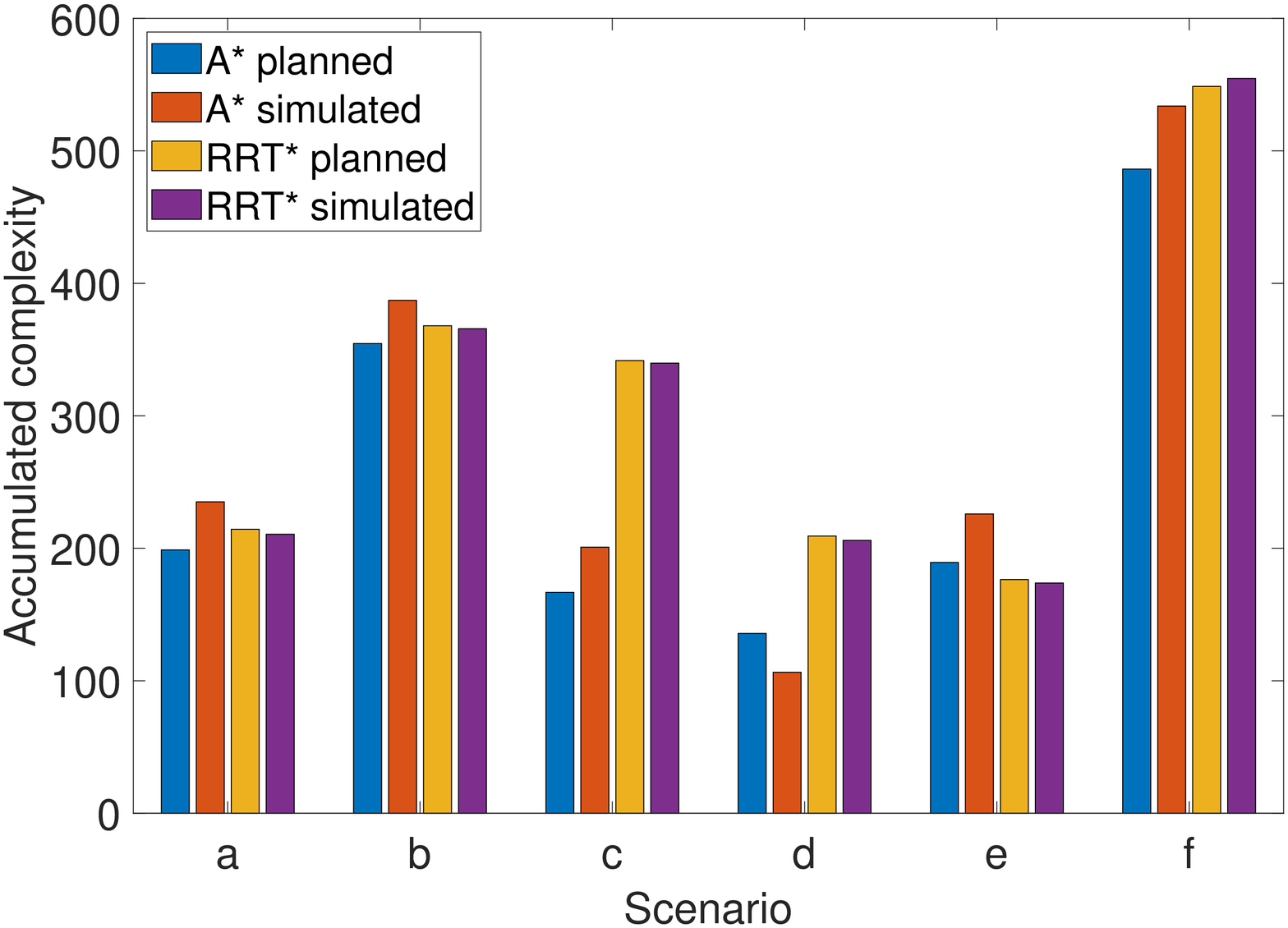}
	\caption{Accumulated complexity for each path planner for the planned (blue, red) and simulated AUV (yellow, purple) paths. Paths are planned with $\alpha = 1000$.}
	\label{fig:path_complexity5}
\end{figure}
\begin{figure}[!h]
	\centering
	\includegraphics[width=.85\columnwidth]{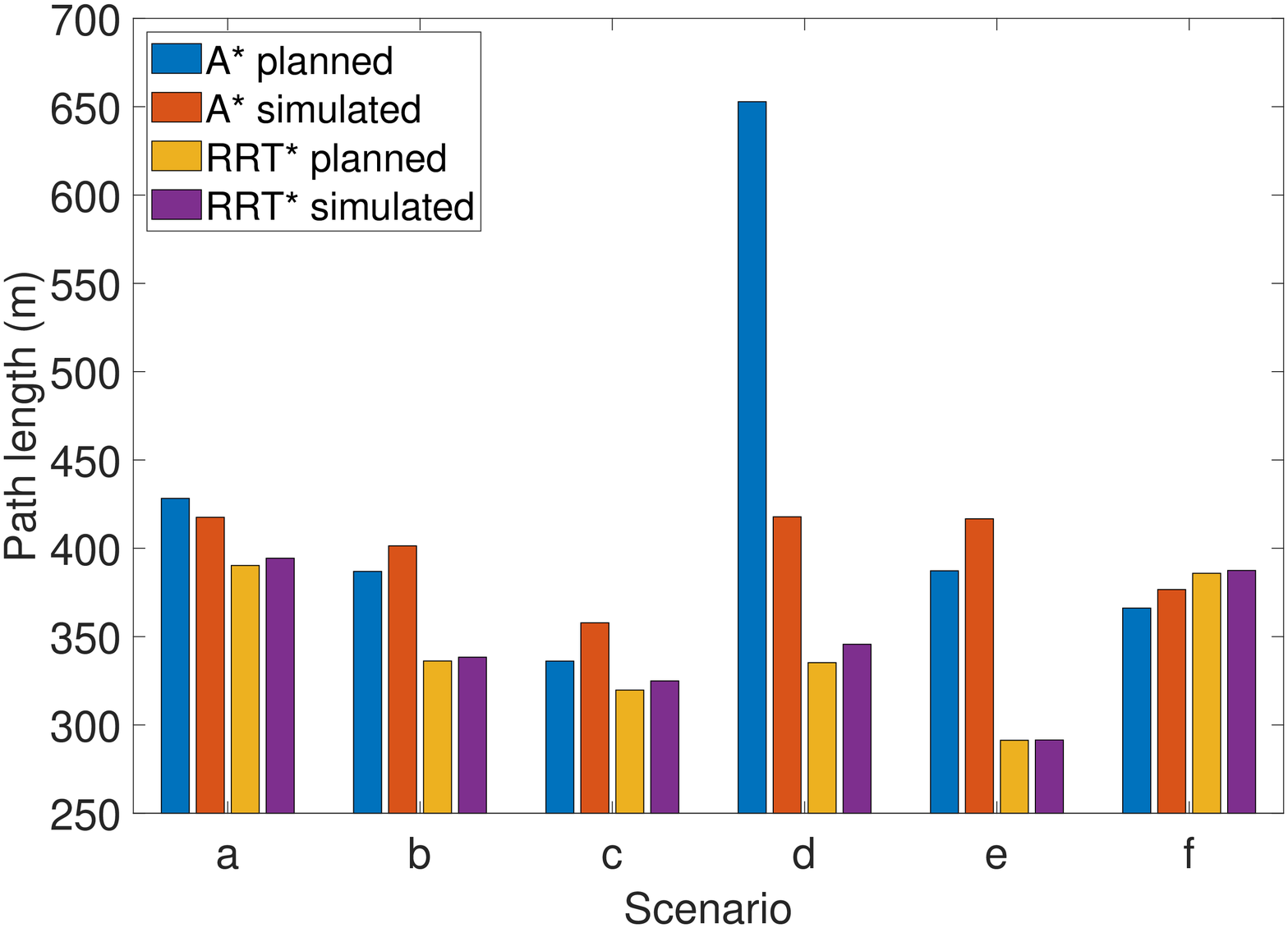}
	\caption{Path length comparison for each path planner for the planned (blue, red) and simulated AUV (yellow, purple) paths. Paths are planned with $\alpha = 1000$.}
	\label{fig:path_length5}
\end{figure}

\end{document}